\documentclass[a4paper,fleqn]{cas-dc}
\usepackage[numbers]{natbib}
\usepackage{amssymb}
\usepackage{amsmath}
\usepackage{float}
\usepackage{verbatim}
\restylefloat{figure}
\floatstyle{plaintop}
\restylefloat{table}
\usepackage{algorithm}
\usepackage{algpseudocode}
\usepackage{graphicx}
\usepackage{subfigure}
\usepackage{subcaption}
\usepackage{longtable}
\usepackage{hyperref}
\usepackage{xcolor}
\usepackage{booktabs} 

\def\tsc#1{\csdef{#1}{\textsc{\lowercase{#1}}\xspace}}
\tsc{WGM}
\tsc{QE}

\begin{document}
\let\WriteBookmarks\relax
\def\floatpagepagefraction{1}
\def\textpagefraction{.001}
\let\printorcid\relax 

\ExplSyntaxOn
\cs_set:Npn \__first_footerline:
{
  \group_begin:
  \small
  \sffamily
  \ifnum\theblind>0\relax
  \else
    { \rmfamily Accepted~ by ~\itshape Knowledge-Based~ Systems, ~DOI: ~\href{https://doi.org/10.1016/j.knosys.2026.116373}{10.1016/j.knosys.2026.116373} }
  \fi
  \group_end:
}
\cs_gset:Nn \__short_title: {}
\ExplSyntaxOff


\shortauthors{Zhongzheng Yuan et al.}

\title[mode = title]{FedSA-GCL: A Semi-Asynchronous Federated Graph Learning Framework with Personalized Aggregation and Cluster-Aware Broadcasting}  

\author[1]{Zhongzheng Yuan}
\ead{genhz@mail.sdu.edu.cn} 
\credit{Conceptualization, Methodology, Software, Data curation, Investigation, Formal analysis, Writing -- original draft}

\author[1]{Lianshuai Guo}
\ead{202437607@mail.sdu.edu.cn} 
\credit{Software, Writing -- review \& editing}

\author[2]{Xunkai Li}
\ead{3120225467@bit.edu.cn} 
\credit{Methodology, Visualization, Supervision, Writing -- review \& editing}

\author[3]{Yinlin Zhu}
\ead{zhuylin27@mail2.sysu.edu.cn} 
\credit{Writing -- review \& editing}

\author[1]{Wenyu Wang}
\cormark[1]
\ead{hochi@sdu.edu.cn} 
\credit{Supervision}

\author[1]{Meixia Qu}
\cormark[1]
\ead{mxqu@sdu.edu.cn} 
\credit{Supervision}


\address[1]{Shandong University, School of Airspace Science and Engineering, Weihai 264209, China}
\address[2]{Beijing Institute of Technology, School of Computer Science and Technology, Beijing 100081, China}
\address[3]{Sun Yat-sen University, School of Computer Science and Engineering, Guangzhou 510275, China}

\begin{abstract}
Federated Graph Learning (FGL) is a distributed learning paradigm that enables collaborative training over large-scale subgraphs located on multiple local systems. However, most existing FGL approaches rely on synchronous communication, which leads to inefficiencies and is often impractical in real-world deployments. Meanwhile, current asynchronous federated learning (AFL) methods are primarily designed for conventional tasks such as image classification and natural language processing, consequently failing to account for the unique topological properties of graph data. Directly applying these methods to graph learning frequently results in semantic drift and representational inconsistency within the global model. To address these challenges, we propose FedSA-GCL, a semi-asynchronous federated framework that leverages both inter-client label distribution divergence and graph topological characteristics through a novel ClusterCast mechanism for efficient training. We evaluate FedSA-GCL on multiple real-world graph datasets using the Louvain and Metis algorithms and conduct comparative analysis against 10 baselines. Extensive experiments demonstrate that our method achieves superior robustness and outstanding efficiency, outperforming the baselines by an average margin of 1.9\% with Louvain and 3.0\% with Metis.
\end{abstract}




\maketitle

\section{Introduction}
\label{introduction}

Graphs are fundamental data structures for modeling complex relational systems across a wide range of domains, including social networks\cite{guo2020deep}, drug discovery\cite{liu2022graphcdr,wang2023gadrp}, and recommendation platforms\cite{he2023simplifying,yu2022graph}. Although graph machine learning has made significant progress in recent years, most existing methods rely on centralized storage of large-scale graph data on a single server. However, with growing concerns about data security and user privacy, this centralized paradigm often encounters significant practical hurdles in real-world scenarios. In practice, graph data is typically distributed across multiple data owners, and cross-site data collection is frequently prohibited by stringent privacy restrictions\cite{fu2022federated}. Consequently, effectively integrating these distributed relational structures to support intelligent prediction and decision-making has emerged as a critical computational requirement for modern knowledge-based systems.

To address this challenge, the concept of Federated Graph Learning (FGL) has emerged, which seamlessly integrates federated learning (FL) with graph learning. FGL enables collaborative model training among distributed graph data holders without requiring raw data exchange, thereby preserving privacy. 
By facilitating collaborative learning across isolated data silos, FGL provides a robust computational framework to fuse localized structural information into global knowledge-based systems.
In FGL, graph data is typically partitioned into local subgraphs held by different clients, such as transportation subnetworks~\cite{zhang2021traffic,li2021spatial}, social communities~\cite{guo2020deep}, or financial chains~\cite{yang2021financial}. A key challenge for FGL is subgraph heterogeneity, which refers to the significant divergence in label distributions and graph structures across clients due to strong node dependencies\cite{baek2023personalized,xie2021federated}, consequently undermining cross-client model generalization. To mitigate the impact of subgraph heterogeneity, recent studies in FGL have begun to explicitly account for the unique structural and distributional characteristics of graph data. For example, several works incorporate topology-aware designs to capture local graph semantics~\cite{zhang2021subgraph,li2024FedGTA}, while others explore personalized modeling or representation-level adaptation to better handle client-specific variations~\cite{tan2022fedproto,baek2023personalized,li2024adafgl}. 

Despite their effectiveness, these methods are still primarily designed under synchronous training protocols, which introduce significant bottlenecks regarding computational and communication efficiency in practical settings~\cite{jiang2022pisces}. Consequently, current FGL research faces two major limitations.

\textbf{L1: Synchronous FGL Suffers from Inefficiency.} Although existing FGL methods have demonstrated promising results~\cite{baek2023personalized,li2024FedGTA,li2024adafgl,zhu2024fedtad}, they are predominantly designed for idealized synchronous settings. However, their reliance on synchronization raises serious concerns for practical deployment. In heterogeneous environments, such as those involving edge devices or clients with limited computing capacity, heterogeneous stragglers can impede the entire training process, leading to prolonged communication rounds and underutilized system resources~\cite{chen2019communication}. Additionally, in real-world scenarios, client participation is inherently sporadic and inconsistent in dynamic network environments. Strictly requiring all selected clients to complete training before aggregation is incompatible with such dynamic participation patterns, rendering synchronous protocols not only inefficient but often infeasible.

\textbf{L2: Asynchronous Settings Amplify Graph Non-IID.} 
Existing asynchronous federated learning (AFL) methods~\cite{xie2019asynchronous,xu2023asynchronous,ma2021fedsa,nguyen2022federated} are primarily developed for domains such as computer vision and natural language processing. These approaches typically assume independent and identically distributed (IID) data, consequently lacking dedicated mechanisms to address the cross-node dependencies and topological heterogeneity intrinsic to graph data.
While pioneering work like SWESALT~\cite{liao2023accelerating} introduces asynchronous communication into FGL, it focuses primarily on training acceleration and explicitly lacks dedicated mechanisms to address the non-IID challenges on graphs.
The inherent non-IID nature of graph data arises from topological heterogeneity and cross-node dependencies. While synchronous FL can mitigate some of this heterogeneity by aggregating updates from all clients simultaneously, asynchronous participation disrupts this balance. The server often aggregates updates from a temporally sparse and biased subset of clients, causing certain structural patterns to be underrepresented. This imbalance amplifies the existing non-IID challenge, further degrading the consistency and generalizability of the global model.

To address the above issues, we propose Federated Semi-Asynchronous Graph Cluster Learning (FedSA-GCL), a novel and scalable optimization strategy that performs effective client clustering based on their node label distribution characteristics. For structurally correlated clients, we assign aggregation weights by jointly considering their topological characteristics and model staleness. Furthermore, leveraging the asynchronous nature of the system, we introduce a ClusterCast mechanism that proactively pushes model updates to inactive clients sharing similar graph structures.

In summary, the main contributions of this paper are:
(1) \textbf{New Paradigm.} To the best of our knowledge, FedSA-GCL represents the first asynchronous FGL study that explicitly addresses the graph non-IID problem. 
(2) \textbf{Novel Method.} We propose FedSA-GCL, a semi-asynchronous federated learning method for graphs that improves cross-client consistency through a cluster-level broadcasting mechanism. This mechanism mitigates the negative impact of stale updates by enabling inactive but semantically similar clients to benefit from peer-side updates.
(3) \textbf{State-of-the-art Performance.} Extensive experiments on eight benchmark graph datasets demonstrate the significant advantages of FedSA-GCL over state-of-the-art baselines in terms of effectiveness (3.0\% accuracy improvement), robustness (3.7\% accuracy improvement), and convergence efficiency (259.4 client trips improvement).

The paper is organized as follows. In Section \ref{Preliminary and Related Work}, we present the preliminaries and related work, including background knowledge on FL and a review of recent advances in FGL and AFL. Section \ref{FedSA-GCL framework} introduces the proposed FedSA-GCL framework, detailing its core components and mechanisms. Section \ref{Experiments} describes the experimental setup and provides a comprehensive analysis and discussion of the results. Finally, Section \ref{Conclusion} concludes the paper with a summary of findings.

\begin{table}[htbp]
\centering
\caption{Summary of main notations}
\label{tab:notations}
\begin{tabular}{l p{0.8\columnwidth}}
\toprule
\textbf{Notation} & \textbf{Description} \\
\midrule
\multicolumn{2}{l}{\textit{Graph Definitions}} \\
$G, G_i$ & Global graph and local subgraph of client $i$ \\
$\mathcal{V}, \mathcal{E}$ & Set of nodes and edges \\
$n, m, f$ & Total number of nodes, edges, and node feature dimension \\
$\mathbf{A}, \mathbf{X}, \mathbf{Y}$ & Adjacency, feature, and label matrices \\
$D_u$ & Degree of node $u$ \\
$N$ & Total number of clients \\
$|\mathcal{Y}|$ & Total number of classes \\
\midrule
\multicolumn{2}{l}{\textit{Federated Learning}} \\
$t$ & Global communication round \\
$K$ & Semi-asynchronous buffer threshold (required updates per round) \\
$\omega$ & Model parameters (subscript $i$ denotes client, superscript $t$ denotes round) \\
$\tau_i, d_{i,t}$ & Last upload round and staleness delay of client $i$ \\
$\eta$ & Learning rate \\
$\mathcal{U}, \mathcal{B}$ & Sets of actively uploaded and broadcast clients \\
\midrule
\multicolumn{2}{l}{\textit{FedSA-GCL Core}} \\
$\hat{\mathbf{y}}_u$ & Soft label vector of node $u$ \\
$\text{SFM}_i$ & Normalized soft label feature matrix of client $i$ \\
$\operatorname{sim}(i, j)$ & Semantic similarity between client $i$ and $j$ \\
$\mathcal{I}_i$ & Semantic cluster assigned to client $i$ \\
$\theta, \lambda, \alpha$ & Similarity threshold, propagation balance, and staleness attenuation \\
$k$ & Number of Non-param LP steps \\
$LSC_i$ & Local smoothness confidence of client $i$ \\
\midrule
\multicolumn{2}{l}{\textit{Convergence Analysis}} \\
$F(\cdot)$ & Objective loss function \\
$L$ & Lipschitz smoothness constant \\
$\sigma_c^2, G^2$ & Intra-cluster variance bound and gradient norm bound \\
\bottomrule
\end{tabular}
\end{table}

\section{Preliminary and Related Work}
\label{Preliminary and Related Work}

\subsection{Problem Formulation}

Due to the lack of publicly available benchmark datasets for FGL, existing works~\cite{zhang2021subgraph, chen2024fedgl, baek2023personalized} typically simulate FGL scenarios by partitioning a single global graph into multiple distributed subgraphs.
Given a global graph $G = (\mathcal{V}, \mathcal{E})$ with $|\mathcal{V}| = n$ nodes and $|\mathcal{E}| = m$ edges, its structure is represented by an adjacency matrix $\mathbf{A} \in \mathbb{R}^ {n \times n}$. Each node $v \in \mathcal{V}$ is associated with a feature vector $\mathbf{x}_v \in \mathbb{R}^f$, forming the feature matrix $\mathbf{X} \in \mathbb{R}^{n \times f}$, and a one-hot label vector $\mathbf{y}_v \in \mathbb{R}^{|\mathcal{Y}|}$, resulting in the label matrix $\mathbf{Y} \in \mathbb{R}^{n \times |\mathcal{Y}|}$, where $|\mathcal{Y}|$ is the number of classes.
To simulate a federated environment, community detection algorithms such as Metis\cite{karypis1998fast} or Louvain\cite{blondel2008fast} are applied to decompose the global graph $G$ into $N$ subgraphs. Each client $i \in \left \{1, \dots, N \right\}$ is assigned a local subgraph $G_i = (\mathcal{V}_i, \mathcal{E}_i)$, along with its corresponding local adjacency matrix $\mathbf{A}_i \in \mathbb{R}^{|\mathcal{V}_i| \times |\mathcal{V}_i|}$, feature matrix $\mathbf{X}_i \in \mathbb{R}^{|\mathcal{V}_i| \times f}$, and label matrix $\mathbf{Y}_i \in \mathbb{R}^{|\mathcal{V}_i| \times |\mathcal{Y}|}$. For clarity and easy reference, the frequently used mathematical notations throughout this paper are summarized in Table~\ref{tab:notations}.

\subsection{Graph Neural Networks}
\label{sec:gnn}
Graph Neural Networks (GNNs) have emerged as a dominant paradigm for learning representations from graph-structured data~\cite{kipf2016semi, hamilton2017inductive, velickovic2018graph}. The core mechanism of most modern GNNs is the neural message-passing framework~\cite{gilmer2017neural}, where each node iteratively updates its own feature representation by aggregating information from its topological neighborhood. Formally, given the initial node features $\mathbf{h}_v^{(0)} = \mathbf{x}_v$, the representation of a node $v \in \mathcal{V}$ at the $l$-th layer, denoted as $\mathbf{h}_v^{(l)}$, is updated via:
\begin{equation}
\begin{split}
    \mathbf{h}_v^{(l)} &= \text{UPDATE}^{(l)} \Big( \mathbf{h}_v^{(l-1)}, \text{AGGREGATE}^{(l)} \Big( \\
    &\quad \{ \mathbf{h}_u^{(l-1)} : u \in \mathcal{N}(v) \} \Big) \Big).
\end{split}
\label{eq:gnn_mp}
\end{equation}

where $\mathcal{N}(v)$ represents the set of neighbors of node $v$, and $\text{AGGREGATE}(\cdot)$ and $\text{UPDATE}(\cdot)$ are differentiable permutation-invariant functions. By stacking multiple layers, GNNs capture high-order structural and attribute information. Representative architectures differ primarily in their design of these two functions. For instance, Graph Attention Networks (GAT)~\cite{velickovic2018graph} leverage self-attention mechanisms to assign varying weights to different neighbors, and Graph Isomorphism Networks (GIN)~\cite{xu2018how} optimize the aggregation process to maximize expressive power.

Among these architectures, GCN~\cite{kipf2016semi} stands out as a foundational baseline that simplifies spectral graph convolutions. In a GCN, the message-passing mechanism is executed across the entire graph structure simultaneously. Given the adjacency matrix $\mathbf{A}$ and the layer input matrix $\mathbf{H}^{(l-1)}$ with $\mathbf{H}^{(0)} = \mathbf{X}$, the layer-wise propagation rule is formulated as:
\begin{equation}
\mathbf{H}^{(l)} = \sigma \left( \tilde{\mathbf{D}}^{-\frac{1}{2}} \tilde{\mathbf{A}} \tilde{\mathbf{D}}^{-\frac{1}{2}} \mathbf{H}^{(l-1)} \mathbf{W}^{(l-1)} \right).
\label{eq:gcn_propagation}
\end{equation}

where $\tilde{\mathbf{A}} = \mathbf{A} + \mathbf{I}$ is the adjacency matrix with added self-loops, and $\tilde{\mathbf{D}}$ is the corresponding degree matrix. The matrix $\mathbf{W}^{(l-1)}$ is a layer-specific trainable weight matrix, and $\sigma(\cdot)$ acts as a non-linear activation function. By intertwining feature transformation with topological aggregation, GCN provides a robust mechanism for extracting node embeddings.

Conventional coupled GNNs suffer from scalability bottlenecks on large-scale graphs due to the neighborhood explosion problem during recursive message passing. To address this limitation, scalable GNNs have been developed, broadly diverging into two paradigms. The first paradigm relies on graph sampling techniques to constrain the neighborhood size. Notable methods include node-wise sampling in GraphSAGE~\cite{hamilton2017inductive}, layer-wise sampling in FastGCN~\cite{chen2018fastgcn}, and subgraph-based partitioning in Cluster-GCN~\cite{chiang2019cluster}.
The second paradigm advocates for a decoupled architecture, which separates topological feature propagation from non-linear neural transformations. For example, SGC~\cite{wu2019simplifying} reduces the multi-layer GCN to a linear model operating on precomputed $k$-step propagated features. Building upon this, SIGN~\cite{frasca2020sign} concatenates features propagated at different scales, while S$^2$GC~\cite{zhu2021simple} and GBP~\cite{chen2020scalable} introduce spectral and bidirectional propagation with moving averages. More recently, GAMLP~\cite{zhang2022graph} incorporates attention mechanisms to adaptively fuse multi-scale topological features. 

While decoupled architectures resolve local scalability bottlenecks, deploying GNNs in a federated environment introduces distinct optimization challenges. Standard federated learning typically addresses non-IID data arising from feature or label skew. In contrast, the neighborhood propagation mechanism in GNNs depends directly on local graph topology, causing isolated client subgraphs to exhibit highly divergent structural patterns. This intrinsic structural heterogeneity renders conventional non-IID solutions inadequate and necessitates specialized optimization strategies designed to explicitly mitigate topological variances across clients.

\subsection{Federated Graph Learning}

FedAvg~\cite{mcmahan2017communication} is the most widely used optimization strategy in FL. It performs global model aggregation by computing a weighted average of the model parameters received from participating clients, where the weights are typically proportional to the size of local training data. In the context of FGL, we integrate GNNs\cite{kipf2016semi} with FedAvg, where the aggregation process at round~$t$ is formulated as:

\begin{equation}
\tilde{\omega}^{t} = \sum_{i=1}^N \frac{n_i}{n} \cdot \omega_i^{t-1}.
\label{eq:FedAvg_aggregation}
\end{equation}

where $\tilde{\omega}^{t}$ is the global model at round $t$, $\omega_i^{t-1}$ is the local model uploaded by client $i$ in the previous round, $n_i$ denotes the number of training nodes on client $i$, and $n = \sum_{i=1}^{N} n_i$ represents the total number of training nodes. The local update on client $i$ during round~$t$ is defined as:

\begin{equation}
\omega_i^{t} = \tilde{\omega}^{t} - \eta \cdot \nabla f(\tilde{\omega}^{t}, G_i(\mathcal{V}_i, \mathcal{E}_i)).
\label{eq:FedAvg_local_update_clean}
\end{equation}

where $\tilde{\omega}^{t}$ is the received global model, $\eta$ is the learning rate, and $\nabla f(\cdot)$ is the gradient of the local loss function computed over the subgraph $G_i$.

Although FedAvg is widely adopted, it often exhibits suboptimal performance under non-IID client data and fails to capture the unique properties of graphs. To address this, FedSage+\cite{zhang2021subgraph} integrates GraphSAGE into the federated setting to reconstruct missing inter-subgraph edges, though it primarily focuses on structural completion. FedProto\cite{tan2022fedproto} tackles label distribution heterogeneity via prototype sharing but overlooks local graph topology. In contrast, FedPub~\cite{baek2023personalized} introduces functional embeddings and sparse aggregation to enable personalized training on heterogeneous subgraphs. Furthermore, FedGTA~\cite{li2024FedGTA} and AdaFGL~\cite{li2024adafgl} incorporate topological confidence and structure-aware personalization strategies, thereby improving both convergence and adaptability. More recently, FedTAD~\cite{zhu2024fedtad} adopts a topology-aware, data-free distillation mechanism to enable structure-sensitive aggregation without raw data sharing. 
To systematically evaluate and deploy these algorithms, platforms like FedGraphNN~\cite{he2021fedgraphnn} and FedGraph~\cite{yao2024fedgraph} have been introduced. FedGraphNN provides a unified benchmark that exposes the limitations of existing models under severe graph heterogeneity, while FedGraph focuses on real-world efficiency, utilizing low-rank communication and homomorphic encryption to evaluate scalable deployment.

However, despite these algorithmic and system-level advancements, most existing FGL frameworks fundamentally rely on synchronous communication protocols, where global aggregation is blocked until all selected clients complete local training. This strict synchronization limits scalability and leads to severe inefficiencies in real-world heterogeneous or unstable environments, leaving the challenges of semi-asynchronous FGL largely unexplored.

\subsection{Asynchronous Federated Learning}

To overcome the inefficiencies of synchronous FL amidst system heterogeneity, AFL has been proposed to enable non-blocking client-server communication~\cite{xu2023asynchronous}. AFL can be broadly categorized based on its aggregation strategy into fully asynchronous and semi-asynchronous approaches. Fully asynchronous methods, such as FedAsync~\cite{xie2019asynchronous} and TWAFL~\cite{chen2019communication}, allow the server to update the global model immediately upon receiving a single client update, thereby improving reactivity but risking optimization instability under non-IID conditions. In contrast, semi-asynchronous methods like FedSA~\cite{ma2021fedsa} and FedBuff~\cite{nguyen2022federated} trigger aggregation only after collecting a predefined number of updates, striking a balance between convergence stability and communication efficiency. TimelyFL~\cite{zhang2023timelyfl} introduces a tier-based scheduling mechanism that adapts local training based on device availability and latency, further enhancing responsiveness in heterogeneous environments.

Although preliminary efforts, such as SWESALT~\cite{liao2023accelerating}, have extended semi-asynchronous strategies to graph data, they lack graph-specific optimizations and overlook the inherent non-IID nature of graph data. Current AFL methods typically assume IID inputs and lack mechanisms to address graph-specific challenges such as topological heterogeneity and cross-node dependencies. As a result, their direct application to graph learning often leads to unstable convergence and limited generalization, underscoring the necessity for graph-aware asynchronous federated algorithms.

\section{FedSA-GCL Framework}
\label{FedSA-GCL framework}

\begin{figure*}[ht]
    \centering
    \includegraphics[width=1\linewidth]{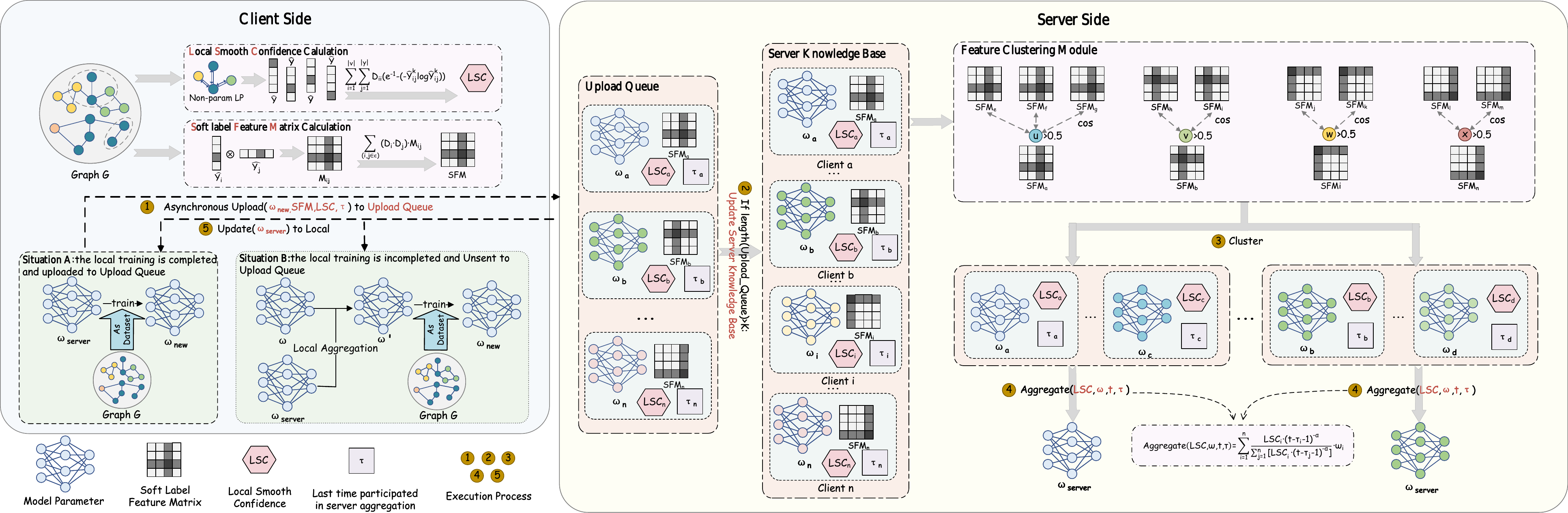}
    \caption{Overview of the FedSA-GCL framework.}
    \label{fig:framework}
\end{figure*}

FedSA-GCL is a semi-asynchronous federated learning framework specifically designed for graph-structured data, where the server initiates aggregation once a predefined number of client updates are received.
As illustrated in Fig.~\ref{fig:framework}, the framework follows a classic client-server architecture, enhanced with several novel designs to address non-IID data and partial communication. Clients are first clustered based on their soft label feature matrices (\textbf{SFM}), and model aggregation is performed by incorporating local smoothness confidence (\textbf{LSC}) and model staleness, thereby promoting better representation alignment across clients. Crucially, our \textbf{ClusterCast mechanism} enables semantically similar but inactive clients to benefit from peer updates, effectively mitigating the model degradation commonly seen in fully asynchronous settings. In Step 4 of Fig.~\ref{fig:framework}, the server performs personalized aggregation over distinct semantic clusters. While the aggregation mechanism (Eq.~\eqref{eq:omega_server}) remains identical across all clusters, it generates distinct customized models tailored to each cluster's unique characteristics. These different models are visually distinguished by different colors in the figure.

(1) The left panel illustrates the client-side workflow, including the generation of SFM, the calculation of LSC, and the handling of asynchronous model updates received during local training via the ClusterCast mechanism.
(2) The right panel presents the server-side workflow, including the similarity-based client clustering process and the aggregation of client models weighted by LSC and model staleness.

In the following sections, we elaborate on each core design component:
Section~\ref{sec:SFM} introduces the construction of SFM and the SFM-based clustering algorithm;
Section~\ref{sec:Aggregation} details the construction of LSC and the staleness-aware aggregation strategy;
Section~\ref{sec:ClusterCast} explains how the ClusterCast mechanism leverages client clustering to support semi-asynchronous propagation.

\subsection{Client Clustering Based on Soft Label Feature Matrix}
\label{sec:SFM}

In classification tasks, a hard label refers to a discrete, one-hot encoded vector that assigns a node to a single category, implying absolute certainty where one class receives full probability and all others receive zero. While straightforward, such rigid assignments fail to reflect the nuanced semantic overlap and relationships between different categories. In contrast, a soft label provides a continuous probability distribution across all classes, offering a more informative representation of the model's confidence and the inherent ambiguity of the data.
Compared with hard labels, soft labels can better capture the uncertainty and inter-class similarity in categorization. Therefore, when modeling client-side label features, we adopt node soft labels as the basis. The soft label of nodes is computed by passing input features through an encoder followed by a softmax activation, as defined in Eq.~\eqref{eq:softlabel}:

\begin{equation}
\label{eq:softlabel}
\hat{\mathbf{Y}}_i = \text{Softmax}\left(f(\tilde{\omega}_i; G_i(\mathcal{V}_i, \mathcal{E}_i))\right).
\end{equation}

where $G_i(\mathcal{V}_i,\mathcal{E}_i)$ denotes the local subgraph of client $i$, and $f(\tilde{\omega}_i; \cdot)$ is the predictive model parameterized by the aggregated weights $\tilde{\omega}_i$.

Since connected nodes often exhibit similarity in both feature distributions and labels, while different clients present distinct label distribution patterns, we construct a Soft Label Feature Matrix (SFM) for adjacent nodes by multiplying their soft label features based on each client's graph data. Typically, nodes closer to the client center (which generally have higher degrees) provide a better representation of the client's label distribution. Therefore, we weigh the adjacent nodes' SFM using node degrees as weights, thereby enhancing the contribution of critical nodes. Finally, the client's overall soft label feature representation is obtained by aggregating all weighted soft label feature matrices. The SFM is computed as follows:

\begin{equation}
\label{eq:SFM}
\text{SFM}_i = \frac{1}{\sum_{(u, v) \in \mathcal{E}_i} D_u D_v} \sum_{(u, v) \in \mathcal{E}_i} \left(D_u \cdot D_v\right) \left( \hat{\mathbf{y}}_u \hat{\mathbf{y}}_v^\top \right).
\end{equation}

where $\mathcal{E}_i$ denotes the set of all edges in client $i$'s subgraph, $u$ and $v$ are two connected nodes, $D_u$ and $D_v$ represent their respective degrees, and $\hat{\mathbf{y}}_u, \hat{\mathbf{y}}_v \in \mathbb{R}^{|\mathcal{Y}|}$ are their soft label column vectors. To ensure that the SFM accurately measures pure label distribution divergence without being skewed by the local graph scale, we introduce the normalization term $\left(\sum_{(u, v) \in \mathcal{E}_i} D_u D_v\right)^{-1}$.

Upon receiving the SFMs from participating clients, the server integrates them into the \textbf{Server Knowledge Base} by replacing outdated historical records with the newly uploaded SFMs and their corresponding model parameters. The server then utilizes this updated repository of stored SFMs to compute pairwise similarities between active clients and all other peers.

The objective of this similarity calculation (Eq.~\eqref{eq:sim}) is to quantitatively measure the alignment of label distributions across clients without exposing their raw local graph data. Specifically, it computes the cosine similarity between the flattened SFMs and applies a predefined threshold $\theta$ to identify a specific set of compatible peers, denoted as $\mathcal{I}_i$, for each client $i$.

\begin{equation}
\label{eq:sim}
\begin{gathered}
\mathcal{I}_i = \{ j \mid \operatorname{sim}(i, j) \geq \theta \} \cup \{i\}, \quad \forall i, j \in \operatorname{Set}(N),\ j \neq i \\
\operatorname{sim}(i, j) = \frac{\sum_{p=1}^{|\mathcal{Y}|^2} \operatorname{SFM}_i^{p} \cdot \operatorname{SFM}_j^{p}}{\sqrt{\sum_{p=1}^{|\mathcal{Y}|^2} \left(\operatorname{SFM}_i^{p}\right)^2} \cdot \sqrt{\sum_{p=1}^{|\mathcal{Y}|^2} \left(\operatorname{SFM}_j^{p}\right)^2}}.
\end{gathered}
\end{equation}

where $\operatorname{SFM}_i^p$ denotes the $p$-th element in the flattened SFM of client $i$, $\mathcal{I}_i$ is the set of clients similar to client $i$, and $\theta$ is the similarity threshold.

This client similarity plays a crucial role in optimizing the overall system topology. Instead of performing a naive global aggregation like standard FedAvg, which often suffers from weight divergence under non-IID graph scenarios, our framework leverages these peer sets to execute targeted, cluster-aware mutual aggregation. This mechanism effectively mitigates negative transfer between clients with highly skewed or disjoint label distributions, thereby promoting more stable and personalized model convergence. Based on these sets, the target clients for subsequent parameter distribution are explicitly determined. The specific calculation process is visually summarized in the \textbf{Feature Clustering Module} of Fig.~\ref{fig:framework}.

It is important to note that while SFM provides informative node features based on soft labels, it inherently omits the topological structures of the graph data. Nevertheless, as empirically validated in our ablation studies (Section~\ref{sec:Interpretability}), this representation remains highly effective for client clustering. To ensure this design does not compromise overall performance, the topological information missed by the SFM at the server side is explicitly compensated for locally by the topology-aware LSC during the subsequent aggregation phase (Section~\ref{sec:Aggregation}).

\subsection{Model Aggregation Based on Local Smoothness Confidence and Staleness}
\label{sec:Aggregation}
To quantify the topological-attribute correlation between nodes, we introduce $k$-step Non-parametric Label Propagation (Non-param LP), which models the relationship between target nodes and their $k$-hop neighbors through iterative updates\cite{gasteiger2018predict}. To capture higher-order structural information, the $k$-step Non-param LP is given by:

\begin{equation}
\label{eq:LP}
\begin{aligned}
\hat{\mathbf{y}}_u^{(k)} &= \lambda \cdot \hat{\mathbf{y}}_u^{(0)} + (1-\lambda) \sum_{v \in \mathcal{N}(u)} \frac{1}{\sqrt{D_u D_v}} \cdot \hat{\mathbf{y}}_v^{(k-1)} \\
&= \left[ \lambda \hat{\mathbf{Y}}_i^{(0)} + (1-\lambda) \tilde{\mathbf{A}}_i \hat{\mathbf{Y}}_i^{(k-1)} \right]_u.
\end{aligned}
\end{equation}

where $\hat{\mathbf{y}}_u^{(0)}$ is the initial soft label vector of node $u$, $\mathcal{N}(u)$ denotes the 1-hop neighbors of node $u$, $D_u$ is the degree of node $u$, and $\lambda \in [0,1]$ controls the propagation balance. In the matrix formulation, $\hat{\mathbf{Y}}_i^{(k)}$ denotes the soft label matrix at step $k$, $\tilde{\mathbf{A}}_i$ is the symmetrically normalized adjacency matrix, and $[\cdot]_u$ extracts the $u$-th row. Importantly, this propagation is not an infinite recursive process, but a fixed, finite $k$-step operation. In practice, it is implemented via efficient sparse matrix multiplications rather than individual node-level iterations. A detailed theoretical complexity analysis demonstrating its linear scalability and negligible overhead is deferred to Section~\ref{sec:efficiency_analysis}.

The entropy of node soft labels quantifies the prediction consistency of local models. Specifically, lower entropy indicates stronger prediction consistency within local graph structures, yielding more reliable learned topological features. Building upon this, the LSC characterizes their consistency with neighborhood predictions\cite{li2024FedGTA}. Furthermore, considering the significant influence of high-degree nodes on LSC, we assign them higher weights. The LSC, which reflects the entropy-based consistency between nodes and their neighbors, is defined as shown in Eq.~\eqref{eq:LSC}:

\begin{equation}
\label{eq:LSC}
\begin{aligned}LSC_i & = \sum_{u \in \mathcal{V}_i} D_u \cdot \left ( e^{-1}-\text{Entropy}(\hat{\mathbf{y}}_u^{(k)})\right ) \\& = \sum_{u \in \mathcal{V}_i} D_u \cdot \left (e^{-1}+\left (\sum_{l = 1}^{|\mathcal{Y}|}\hat{y}_{u,l}^{(k)}\cdot \log \hat{y}_{u,l}^{(k)} \right )\right ). \end{aligned}
\end{equation}

where $\mathcal{V}_i$ is the set of nodes in client $i$, $|\mathcal{Y}|$ is the soft label dimensionality (number of classes), $l$ is the class index, and $D_u$ is the degree of node $u$.

Importantly, while the formulation of LSC builds upon entropy-based metrics in existing literature, its adoption here is driven by its excellent capability to extract topological features and evaluate local model quality. This makes it highly suitable for our scenario, as it explicitly compensates for the deep topology missed by the SFM clustering. 
However, directly applying such a spatial metric in an AFL framework is insufficient. In AFL frameworks, clients operate with varying computational speeds and communication delays. Consequently, some clients may train locally based on highly outdated versions of the global model. This delay between the current server round and the historical model version upon which a client's update is based is defined as model version staleness~\cite{xie2019asynchronous}. If left unaddressed, aggregating stale updates can introduce gradient conflicts and destabilize global convergence. 
To mitigate this issue, we employ a staleness metric to dynamically penalize outdated updates and adjust client model aggregation weights. The staleness-aware aggregation weight is formulated as:

\begin{equation}
\label{eq:omega_server}
\tilde{\omega}_i =\sum_{j \in \mathcal{I}_i}\frac{LSC_j\cdot (d_{j,t})^{-\alpha }}{\sum_{c \in \mathcal{I}_i} \left (LSC_c\cdot (d_{c,t})^{- \alpha } \right ) } \cdot \omega_j.
\end{equation}

where $t$ is the current server round, $\tau_j$ is the last upload round of client $j \in \mathcal{I}_i$, $d_{j,t} = \max(1, t - \tau_j)$ is its staleness delay, $\omega_j$ is its model, and $\alpha$ is a staleness attenuation coefficient. 

Intuitively, the term $d_{j,t}$ explicitly quantifies the staleness gap, defined as the number of global updates that occurred during the client's local training. By applying the negative exponent $-\alpha$, Eq.~\eqref{eq:omega_server} functions as a mathematical decay mechanism: the staler a client's update, the strictly smaller its assigned weight. This design effectively suppresses the negative impact of stale models, ensuring that the global aggregation prioritizes both high-quality (high LSC) and fresh (low staleness) local graph representations.

Therefore, our core algorithmic contribution in this module lies in the mathematical coupling of this spatial confidence ($LSC$) with the temporal penalty ($(d_{j,t})^{-\alpha}$). Eq.~\eqref{eq:omega_server} thereby establishes a cohesive spatio-temporal filtration mechanism: the LSC dimension acts as a spatial filter to down-weight topologically noisy updates, while the staleness dimension acts as a temporal filter to suppress obsolete models. The necessity of this specific combination is empirically validated by our ablation studies (Section~\ref{sec:Interpretability}), where relying solely on LSC without staleness awareness leads to severe convergence instability. Furthermore, as formally detailed in our convergence analysis (Section~\ref{sec:theory_convergence}), this spatio-temporal coupling is mathematically essential to guarantee the global convergence of the asynchronous system.

\subsection{Semi-Asynchronous ClusterCast Broadcasting}
\label{sec:ClusterCast}

\begin{figure*}[htbp]
    \centering
    \subfigure[FedBuff Communication Diagram\label{fig:comm_diagram_sub1}]{
        \includegraphics[width=0.45\textwidth]{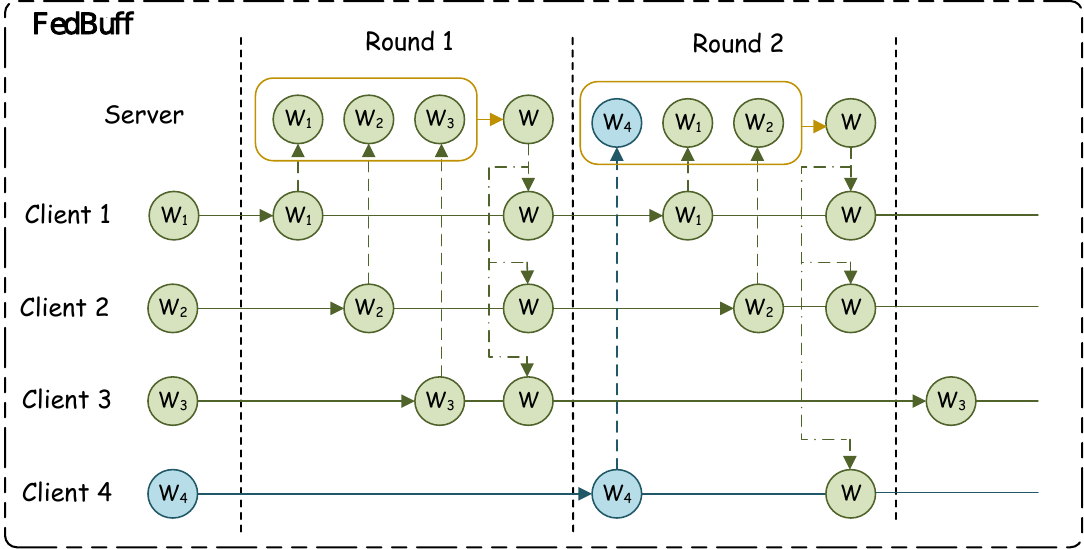}
    }
    \subfigure[FedSA-GCL Communication Diagram\label{fig:comm_diagram_sub2}]{
        \includegraphics[width=0.45\textwidth]{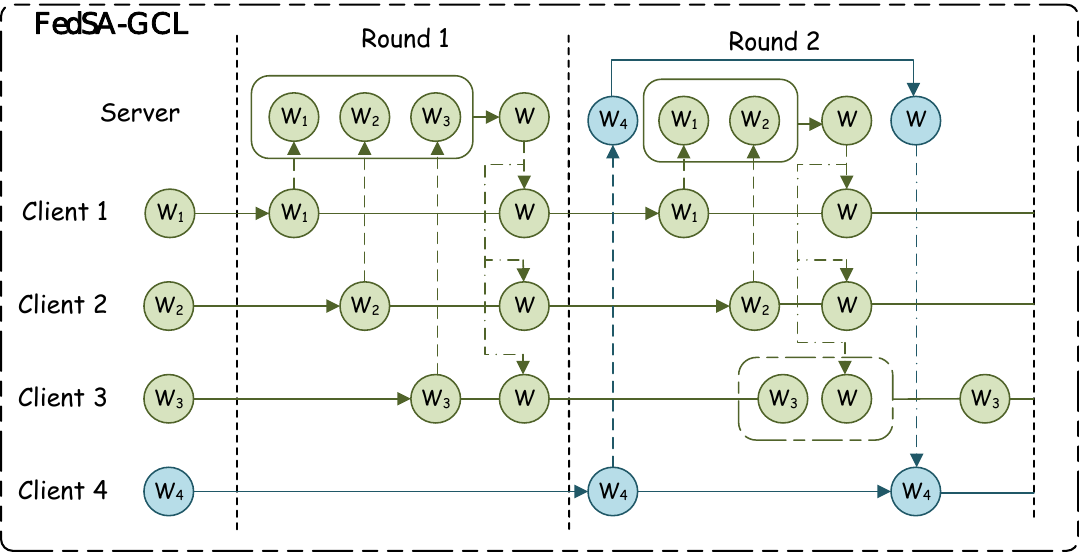}
    }
    \includegraphics[width=0.92\textwidth]{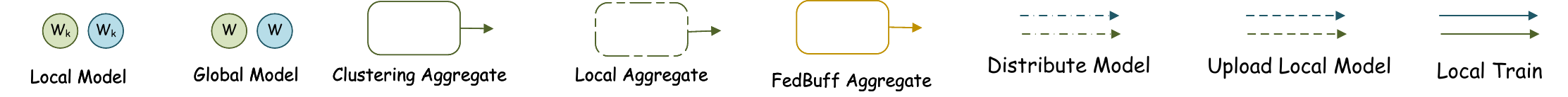}
    \caption{Comparison of communication mechanisms.}
    \label{fig:comm_diagram}
\end{figure*}

A key challenge in semi-asynchronous federated learning is maintaining a timely and coherent information flow under partial client participation. Existing methods like FedBuff\cite{nguyen2022federated} trigger aggregation upon receiving a fixed number of client updates but only send the aggregated model to the clients that participated in that specific round. As illustrated by client 3 in round 2 of Fig.~\ref{fig:comm_diagram_sub1}, this limitation can cause semantically similar but inactive clients to miss out on valuable updates, ultimately leading to slower convergence and increased inter-client drift. To address this issue, we propose \textbf{ClusterCast}, a novel broadcasting mechanism designed to enhance cluster-level cross-client consistency and training efficiency.

ClusterCast proactively pushes model updates to both currently uploaded clients and inactive ones that share similar graph structures. As shown in Fig.~\ref{fig:comm_diagram_sub2}, this allows semantically similar clients to benefit from timely updates even without participating in the current round. This design offers two key advantages: (1) it intensifies cross-client information sharing among semantically similar clients; and (2) it substantially accelerates convergence by reducing parameter divergence.

Specifically, during each communication round, the server aggregates a personalized model $\omega_i$ for each uploaded client $i$ based on its cluster $\mathcal{I}_i$ (Eq.~\eqref{eq:sim}) and optimized model parameters (Eq.~\eqref{eq:omega_server}). The resulting $\omega_i$ is then distributed not only to client $i$, but also to all $j \in \mathcal{I}_i \setminus \mathcal{U}$, where $\mathcal{U}$ is the set of currently uploaded clients. This enables \textbf{semantically similar peers} to benefit from peer updates even during periods of local inactivity.

To ensure training continuity, clients receiving updates during local training do not overwrite their models immediately. Instead, they cache the incoming model. Before initiating the next round of local training, they perform a \textbf{confidence-aware aggregation} between the downloaded server model and their current local model, as defined in Eq.~\eqref{eq:omega_local}:
\begin{equation}
\label{eq:omega_local}
\begin{split}
    \omega_{i}^{agg} &= \frac{\tilde{LSC}_i}{\tilde{LSC}_i + LSC_{i}} \cdot \tilde{\omega}_i + \frac{LSC_{i}}{\tilde{LSC}_i + LSC_{i}} \cdot \omega_{i}, \\
    &\quad \text{where } \tilde{LSC}_i = \sum_{j \in \mathcal{I}_i} LSC_j.
\end{split}
\end{equation}

where $\tilde{\omega}_i$ is the downloaded server model, and $\omega_{i}$ is the current local model at client $i$. The resulting aggregated model $\omega_{i}^{agg}$ (denoted simply as $\omega$ in Algorithm~\ref{alg:client}) serves as the initialization point for the subsequent local training phase, which ultimately produces the newly updated model $\omega_i'$. $\tilde{LSC}_i$ and $LSC_{i}$ represent the aggregated and local LSC values, respectively.

\begin{algorithm}[H]
\caption{FedSA-GCL: Server Procedure}
\label{alg:server}
\begin{algorithmic}[1]
\State \textbf{Initialize:} $t \gets 0$, $\mathcal{U} \gets \emptyset$ \Comment{Global round, uploaded client set}
\State $\texttt{UploadQueue} \gets \texttt{Queue()}$ \Comment{Thread-safe queue for uploaded clients}

\While{$\text{UploadQueue.qsize()} \ge K$}
    \State $t \gets t + 1$, $\mathcal{U} \gets \emptyset$
    \While{$\text{UploadQueue.qsize()} \ne 0$}
        \State Receive $(LSC_i, SFM_i, \omega_i, \tau_i, i)$ from \texttt{UploadQueue.get()}
        \State Update \texttt{ServerKnowledgeBase} with $(LSC_i, SFM_i, \omega_i, \tau_i)$
        \State $\mathcal{U} \gets \mathcal{U} \cup \{i\}$
    \EndWhile
    \For{each client $i \in \mathcal{U}$} \Comment{Distribute personalized model to each uploaded client}
        \State Compute aggregation set $\mathcal{I}_i$ via Eq.~\eqref{eq:sim}
        \State Compute aggregated model $\tilde{\omega}_i$ via Eq.~\eqref{eq:omega_server}
        \State Send $(\tilde{\omega}_i, t, \texttt{null})$ to $\text{DownloadQueue}_i$
        \State $\tilde{LSC}_i \gets \sum\nolimits_{c \in \mathcal{I}_i} LSC_c$
        \For{each client $s \in \mathcal{I}_i \setminus \mathcal{U}$} \Comment{Broadcast to semantically similar non-uploaded clients}
            \State Send $(\tilde{\omega}_i, t, \tilde{LSC}_i)$ to $\text{DownloadQueue}_s$
        \EndFor
    \EndFor
\EndWhile
\end{algorithmic}
\end{algorithm}

\begin{algorithm}[ht]
\caption{FedSA-GCL: Client Procedure}
\label{alg:client}
\begin{algorithmic}[1]
\State \textbf{Initialize:} $\tau \gets 0$, $\texttt{DownloadQueue} \gets \texttt{Queue()}$, $\texttt{update} \gets \texttt{null}$ 

\While{$\texttt{DownloadQueue.qsize()} \ne 0$}
    \State $\texttt{update} \gets \texttt{DownloadQueue.get()}$ \Comment{Retain only the latest update}
\EndWhile

\If{$\texttt{update} \ne \texttt{null}$}
    \State Extract $(\tilde{\omega}, \tau_{new}, \tilde{LSC})$ from $\texttt{update}$
    \State $\tau \gets \tau_{new}$ \Comment{Temporarily cache global round $t$ as local timestamp $\tau$}
    \If{$\tilde{LSC} = \texttt{null}$}
        \State $\omega \gets \tilde{\omega}$
    \Else
        \State Compute $\omega$ via Eq.~\eqref{eq:omega_local} 
    \EndIf
    \State $\texttt{update} \gets \texttt{null}$ \Comment{Avoid repetitive aggregation}
\EndIf

\State Train model $\omega'$ using local data and $\omega$
\State Compute soft label via Eq.~\eqref{eq:softlabel}, $SFM$ via Eq.~\eqref{eq:SFM}
\State Compute Non-param LP via Eq.~\eqref{eq:LP}, $LSC$ via Eq.~\eqref{eq:LSC}
\State Send $(\omega', \tau, SFM, LSC, \text{client ID})$ to \texttt{UploadQueue}
\end{algorithmic}
\end{algorithm}

The ClusterCast mechanism is seamlessly integrated into the FedSA-GCL protocol through coordinated client and server routines, as detailed in Algorithm~\ref{alg:server} and Algorithm~\ref{alg:client}. To ensure clarity and robustness under the semi-asynchronous setting, we highlight several key conventions:
(1) each client maintains a thread-safe download queue and retains only the most recent update, discarding stale ones before training;
(2) clients apply server updates only before starting a new local training cycle to avoid conflicts between overlapping computation phases;
(3) the server broadcasts its global round $t$, which the client temporarily caches as the local timestamp $\tau$. After training, the client uploads its new model $\omega'$ alongside this $\tau$. The server then records this value as $\tau_i$ (the client's last historical version) to strictly compute the staleness delay $d_{i,t}$ for aggregation weighting;
(4) for each uploaded client, semantically similar but non-uploaded clients are grouped into a broadcast set to receive proactive model pushes;
(5) local aggregation is selectively performed only when the server update includes structural confidence information ($\tilde{LSC}$), enabling consistency-aware adaptation.

On the server side (Algorithm~\ref{alg:server}), lines 13-15 identify non-uploaded but semantically similar clients and proactively broadcast the newly aggregated models to them. Correspondingly, on the client side (Algorithm~\ref{alg:client}), lines 2-5 process all incoming updates and retain only the latest one for local training. These coordinated routines enable timely model propagation and alignment within structurally coherent client groups.

\textbf{ClusterCast is the core mechanism that enables FedSA-GCL to operate effectively under semi-asynchronous conditions.} By ensuring model alignment is maintained across similar clients regardless of their upload frequency, it significantly mitigates stale update problems and accelerates convergence, particularly in heterogeneous federated graph scenarios.

\begin{figure}[ht]
        \centering
        \includegraphics[width=1\linewidth]{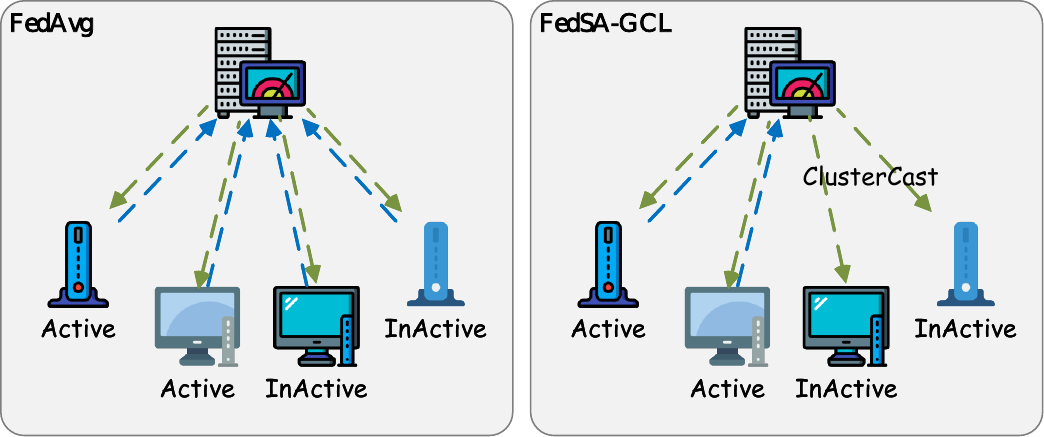}
        \caption{Comparison of per-round communication overhead.}
        \label{fig:comm_overhead_comparison}
\end{figure}

\textbf{Communication Overhead of ClusterCast.} To evaluate communication overhead, we compare the data transmission volume of FedSA-GCL against FedAvg (Fig.~\ref{fig:comm_overhead_comparison}). FedAvg relies on strictly bidirectional communication, requiring both upstream and downstream transmissions for every participating client. Conversely, FedSA-GCL mitigates this critical bottleneck through ClusterCast. As shown in Fig.~\ref{fig:comm_overhead_comparison}, the server collects uploads only from active clients without waiting for stragglers or temporarily disconnected devices. For inactive clients within the same semantic cluster, ClusterCast performs a unidirectional downstream broadcast. These clients receive the updated model to correct local semantic drift without incurring costly upstream transmissions.
When evaluated under traditional round-based metrics requiring all clients to participate, FedSA-GCL significantly lowers the per-round \textbf{upstream} communication burden compared to standard synchronous aggregation. Although FedSA-GCL requires transmitting additional information (LSC, SFM), this structural overhead is negligible because the dense model parameters fundamentally dominate the overall transmission volume. However, it is worth noting that when evaluated under the metric of absolute communication bytes rather than client trips, FedSA-GCL inevitably incurs a higher downstream overhead than the baseline methods due to the proactive ClusterCast broadcasts. This asymmetric design represents a deliberate and highly cost-effective trade-off: we leverage the relatively abundant and cheaper downstream bandwidth to drastically reduce the total number of required client trips, prioritizing rapid global convergence and system stability over minimizing downstream payload.

\textbf{Dynamic Client Participation.} In practical asynchronous federated environments, edge devices may occasionally become inactive or regain connectivity due to unstable networks. FedSA-GCL sustains robust training under these dynamics through three coordinated mechanisms. First, it employs \textbf{threshold-triggered aggregation}. The server initiates a global update upon receiving a predefined number of local updates, naturally bypassing inactive clients without hindering global training progress. Second, the framework facilitates \textbf{automatic recovery via ClusterCast}. The server does not explicitly monitor or target reconnected clients. Instead, it routinely broadcasts aggregated models to all clients within corresponding semantic clusters. A reconnected client passively receives the next scheduled broadcast, effectively overwriting its stale models without requiring a targeted re-activation protocol. Third, a \textbf{staleness penalty serves as a safeguard}. If a previously inactive client uploads an outdated update computed prior to its disconnection, the staleness-aware aggregation (Eq.~\eqref{eq:omega_server}) discounts its contribution, preventing semantic divergence. Finally, the efficacy of this recovery is contingent upon the prior initialization of the client's semantic profile in the \textbf{Server Knowledge Base}. If a device becomes inactive before its initial SFM upload, the server lacks the requisite context to determine its cluster membership, effectively excluding it from the ClusterCast cycle.

\section{Privacy Analysis}
\label{sec:theory_privacy}

FGL inherently preserves data locality by keeping raw node features and edge connections on local devices. Here, we analyze the privacy guarantees and limitations of FedSA-GCL, alongside its compatibility with existing security protocols.

\textbf{Node-Level Privacy.} The mathematical construction of the shared metadata (SFM and LSC) provides a protective compression mechanism for local graph details. Specifically, the SFM condenses the complex adjacency structure and soft predictions into a fixed-size $|\mathcal{Y}| \times |\mathcal{Y}|$ matrix, while the LSC aggregates graph-wide structural entropy into a single scalar. Since the number of classes $|\mathcal{Y}|$ is typically orders of magnitude smaller than the number of local nodes, reverse-engineering individual node features or edge connectivity from these highly compressed representations is computationally prohibitive. Consequently, these compressed representations reduce the direct exposure of node-level information.

\textbf{Class Distribution Leakage.} While individual node data remains secure, we acknowledge a theoretical risk regarding macroscopic statistics. Because the SFM is designed to capture semantic affinities for client clustering, it inherently encodes the overall class distribution of the local dataset. Consequently, continuously transmitting these matrices across communication rounds may potentially reveal a client's global label proportions to an honest-but-curious server or other network participants.

\textbf{Privacy Compatibility.} To address the aforementioned leakage risk, the architecture of FedSA-GCL naturally accommodates advanced privacy-preserving technologies without requiring structural modifications. Established techniques, such as Local Differential Privacy (LDP) or Secure Multi-Party Computation (SMPC), can be seamlessly integrated as pluggable modules to perturb or encrypt the SFM prior to transmission. This modular plug-and-play capability ensures that practical deployments can mitigate global label distribution leakage without fundamentally compromising the underlying semantic clustering mechanism.

\section{Convergence Analysis}
\label{sec:theory_convergence}

To theoretically validate the optimization stability of FedSA-GCL, we provide a formal convergence analysis of our semi-asynchronous aggregation mechanism. Given our cluster-aware aggregation design, we analyze the convergence of the cluster-specific objective function $F_c(\omega) = \sum_{i \in \mathcal{I}_c} p_i F_i(\omega)$ under standard non-convex assumptions, where $p_i$ denotes the inherent data proportion of client $i$.

\subsection{Assumptions}
Our analysis relies on the following standard assumptions commonly adopted in the federated optimization literature:

\begin{itemize}
    \item \textbf{Assumption 1 (Smoothness).} The cluster objective $F_c(\omega)$ is continuously differentiable and $L$-smooth. For any models $\nu$ and $\omega$, it satisfies:
    \begin{equation}
        F_c(\nu) \leq F_c(\omega) + \langle\nabla F_c(\omega), \nu - \omega\rangle + \frac{L}{2}\|\nu - \omega\|^2.
    \end{equation}
    \item \textbf{Assumption 2 (Gradient Diversity).} For any target cluster $\mathcal{I}_c$, the variance of stochastic gradients is bounded by an intra-cluster variance $\sigma_c^2$, such that for any client $i \in \mathcal{I}_c$:
    \begin{equation}
        \mathbb{E}[\|\nabla F_i(\omega) - \nabla F_c(\omega)\|^2] \leq \sigma_c^2.
    \end{equation}
    \item \textbf{Assumption 3 (Bounded Norm and Delay).} The expected squared norm of the stochastic gradients is uniformly bounded by a constant $G^2$, such that $\mathbb{E}[\|\nabla F_i(\omega)\|^2] \leq G^2$. Furthermore, the maximum staleness delay for any client update during the asynchronous process is strictly bounded by a constant $\tau_{max}$, such that $d_{i,t} \leq \tau_{max}$.
    \item \textbf{Assumption 4 (Aggregation Bias).} Due to partial client participation ($\mathcal{U}_t$) and dynamic non-uniform weighting ($q_{i,t} \neq p_i$), the expected pseudo-gradient inherently deviates from the exact cluster gradient. We assume this aggregation bias is bounded by an absolute constant $\epsilon_w^2$ that is independent of the training dynamics, such that $\mathbb{E}[\|\sum_{i \in \mathcal{U}_t} q_{i,t} \nabla F_i(\omega) - \nabla F_c(\omega)\|^2] \leq \epsilon_w^2$.
\end{itemize}

\subsection{Update Formulation}
Let $\tilde{\omega}^t$ denote the server model at global round $t$. During a semi-asynchronous update, the server aggregates models from a subset of uploaded clients $\mathcal{U}_t$. The effective global update is formulated as a pseudo-gradient descent step:

\begin{equation} \label{eq:update_rule}
    \tilde{\omega}^{t+1} = \tilde{\omega}^t - \eta \sum_{i \in \mathcal{U}_t} q_{i,t} \nabla F_i(\tilde{\omega}^{t - d_{i,t}}).
\end{equation}

where $\eta$ is a constant learning rate, $d_{i,t} = \max(1, t - \tau_i)$ represents the staleness delay of client $i$, and $q_{i,t}$ is the normalized staleness-aware aggregation weight defined in Eq.~\eqref{eq:omega_server}. By construction, $q_{i,t} \propto \text{LSC}_i \cdot (d_{i,t})^{-\alpha}$.

\subsection{Main Results}
\textbf{Theorem 1.} Under Assumptions 1 to 4, and assuming a constant learning rate $\eta \leq \frac{1}{4L}$, the sequence of outputs generated by FedSA-GCL after $T$ global rounds converges to a stationary error neighborhood, satisfying the following ergodic bound:

\begin{equation}
\label{eq:theorem_bound}
\begin{split}
    \frac{1}{T} \sum_{t=0}^{T-1} \mathbb{E}[\|\nabla F_c(\tilde{\omega}^t)\|^2] 
    &\leq \mathcal{O} \biggl( \frac{F_c(\tilde{\omega}^0) - F_c(\omega^*)}{\eta T} \\
    &\quad + \eta L \sigma_{c, \text{eff}}^2 + \epsilon_w^2 + \eta^2 L^2 G^2 \tau_{\text{eff}}^2 \biggr).
\end{split}
\end{equation}

where $\omega^*$ is the optimal model, and $\epsilon_w^2$ limits the irreducible aggregation bias. Crucially, to ensure a valid ergodic bound across all rounds, the effective variance is bounded by the supremum $\sigma_{c, \text{eff}}^2 = \sup_t \sum_{i \in \mathcal{U}_t} q_{i,t}^2 \sigma_c^2$, and the cumulative asynchronous drift is tightly bounded by the worst-case effective staleness $\tau_{\text{eff}} = \sup_t \sum_{i \in \mathcal{U}_t} q_{i,t} d_{i,t}$.

\textbf{Remark 1 (Robustness to Clustering Noise).} A critical concern regarding clustered FL is the reliance on initial clustering accuracy. If the initial SFM clustering is suboptimal, misclustered clients may introduce large deviations, potentially violating the bounded intra-cluster variance $\sigma_c^2$. However, our theoretical bound in Eq.~\eqref{eq:theorem_bound} demonstrates that FedSA-GCL remains convergent even under such initial perturbations. While Assumption 4 mathematically limits the absolute worst-case aggregation bias, in practice, the topological confidence filtration via LSC actively minimizes this variance. For any misclustered client, the semantic mismatch between the cluster model and its local data inevitably leads to high prediction uncertainty. According to Eq.~\eqref{eq:LSC}, this elevated structural entropy naturally decays its LSC score. Consequently, through Eq.~\eqref{eq:omega_server}, these topology-induced noisy updates are autonomously down-weighted during aggregation. Thus, the effective intra-cluster variance $\sigma_{c, \text{eff}}^2$ is dynamically suppressed, ensuring that FedSA-GCL does not require perfect initial SFM clustering for guaranteed convergence.

\textbf{Proof Sketch.} Let $\tilde{g}_t = \sum_{i \in \mathcal{U}_t} q_{i,t} \nabla F_i(\tilde{\omega}^{t - d_{i,t}})$. Using the Lipschitz smoothness property (Assumption 1) and the update rule in Eq.~\eqref{eq:update_rule}, taking the total expectation conditioned on the model at round $t$ yields the one-step descent inequality:
\begin{equation} \label{eq:proof_step1}
    \mathbb{E}[F_c(\tilde{\omega}^{t+1})] \leq F_c(\tilde{\omega}^t) - \eta \langle\nabla F_c(\tilde{\omega}^t), \mathbb{E}[\tilde{g}_t]\rangle + \frac{\eta^2 L}{2} \mathbb{E}[\|\tilde{g}_t\|^2].
\end{equation}

To bound the inner product term, we decompose the stale gradient $\mathbb{E}[\tilde{g}_t]$ into the true current gradient $\nabla F_c(\tilde{\omega}^t)$, an aggregation bias vector, and a staleness deviation vector. Applying Young's inequality, the deviation is bounded by:
\begin{equation}
\label{eq:proof_step2}
\begin{split}
    -\langle\nabla F_c(\tilde{\omega}^t), \mathbb{E}[\tilde{g}_t]\rangle 
    &\leq -\frac{1}{2}\|\nabla F_c(\tilde{\omega}^t)\|^2 + \|\Delta_t^{\text{bias}}\|^2 \\
    &\quad + \frac{L^2}{2} \mathbb{E}\Biggl[ \biggl( \sum_{i \in \mathcal{U}_t} q_{i,t} \|\tilde{\omega}^{t} - \tilde{\omega}^{t-d_{i,t}}\| \biggr)^2 \Biggr].
\end{split}
\end{equation}

The rightmost term in Eq.~\eqref{eq:proof_step2} represents the cumulative asynchronous drift. Applying Jensen's inequality and Assumption 3, this drift scales proportionally with $\eta^2 ( \sum_{i \in \mathcal{U}_t} q_{i,t}$ $d_{i,t} )^2 G^2$, which exactly corresponds to the squared effective staleness weighted by the learning rate, i.e., $\eta^2 G^2 \tau_{\text{eff}}^2$. 

Crucially, the parameter $q_{i,t}$ theoretically governs this bound. In standard AFL, this error grows quadratically with the maximum delay $\tau_{max}$. In contrast, our aggregation mechanism explicitly enforces the spatio-temporal coupling $q_{i,t} \propto \text{LSC}_i \cdot (d_{i,t})^{-\alpha}$. This dual-constraint yields two fundamental optimization properties: 
(1) \textbf{Bounded Asynchronous Drift:} The effective staleness is defined as $\tau_{\text{eff}} = \sup_t \sum_{i \in \mathcal{U}_t} q_{i,t} d_{i,t}$. By enforcing $q_{i,t} \propto d_{i,t}^{-\alpha}$, the inner summation term becomes proportional to $d_{i,t}^{1-\alpha}$. Mathematically, as long as the attenuation coefficient is set to $\alpha \geq 1$, the term $d_{i,t}^{1-\alpha}$ is strictly bounded by $\mathcal{O}(1)$, regardless of how infinitely large the actual delay $d_{i,t}$ grows in an asynchronous system. This rigorously prevents the effective staleness $\tau_{\text{eff}}$ from diverging, ensuring stable optimization even under extreme asynchrony. 
(2) \textbf{Variance Reduction:} As analyzed in Remark 1, incorporating $\text{LSC}_i$ assigns inherently lower weights to subgraphs with high structural entropy, thereby actively filtering out topology-induced noise. This mechanism restricts the weight distribution $q_{i,t}^2$, maintaining a tightly bounded effective intra-cluster variance $\sigma_{c, \text{eff}}^2 = \sup_t \sum_{i \in \mathcal{U}_t} q_{i,t}^2 \sigma_c^2$.

Finally, we expand the second moment $\mathbb{E}[\|\tilde{g}_t\|^2] \leq 2\|\nabla F_c(\tilde{\omega}^t)\|^2 + \mathcal{O}(\tau_{\text{eff}}^2 + \epsilon_w^2 + \sigma_{c, \text{eff}}^2)$. Substituting this back into Eq.~\eqref{eq:proof_step1} and restricting $\eta \leq \frac{1}{4L}$ ensures the coefficient of $\|\nabla F_c(\tilde{\omega}^t)\|^2$ is strictly upper-bounded by $-\frac{\eta}{4}$. Telescoping the sum from $t=0$ to $T-1$ derives the final convergence bound in Eq.~\eqref{eq:theorem_bound}. This theoretical formulation explicitly proves that semantic clustering paired with LSC restricts topology-induced variance, while staleness penalties jointly guarantee bounded drift.

\section{Experiments}
\label{Experiments}

In this section, we conduct extensive experiments to validate the effectiveness of our proposed method. We first introduce the eight graph datasets used for training, describe the experimental setup in detail, and explain the partitioning of data into 20 clients using the Metis and Louvain community detection algorithms. FGL is then performed on ten baseline methods. Subsequently, we aim to answer the following research questions:
\textbf{Q1:} In asynchronous scenarios, can our method achieve superior predictive performance compared to state-of-the-art baselines on non-IID data?
\textbf{Q2:} What are the primary sources of the performance improvement in our method?
\textbf{Q3:} How sensitive is our proposed framework to critical hyperparameters during the training process?
\textbf{Q4:} How robust is our method in asynchronous federated graph learning settings?
\textbf{Q5:} Does our method demonstrate higher computational efficiency under asynchronous execution?

\subsection{Experiments Setup}

\textbf{Dataset.} To comprehensively evaluate the performance of our method against other baselines and ensure its scalability across different graph scales, we select eight graph datasets. Our experiments are conducted on three small-scale citation networks (Cora, CiteSeer, PubMed)~\cite{yang2016revisiting}, two medium-scale user-item datasets (Amazon Computers, Amazon Photo), two medium-scale co-authorship networks (Coauthor CS, Coauthor Physics)~\cite{shchur2018pitfalls}, and one large-scale graph (ogbn-arxiv). Detailed descriptions of these datasets and the proportions of the train/validation/test splits are presented in Table~\ref{table:datasetSplit}.

\begin{figure*}
    \centering
    \includegraphics[width=1\linewidth]{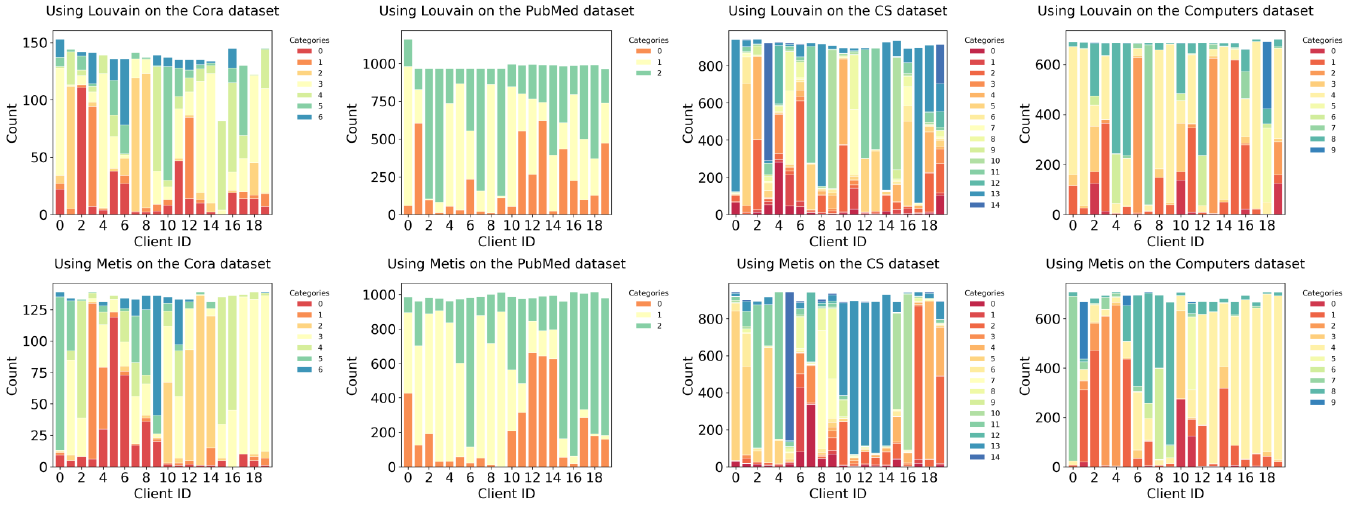}
    \caption{Client-wise label distributions for selected datasets split via Metis and Louvain, demonstrating the resulting non-IID data heterogeneity.}
    \label{fig:datasetSplit}
\end{figure*}

To simulate realistic FGL scenarios, we apply two community detection algorithms, Metis\cite{karypis1998fast} and Louvain\cite{blondel2008fast}, to split each dataset into 20 clients. These splits are used for federated graph training. For several representative datasets and both community detection algorithms, the label distributions across clients are illustrated in Fig.~\ref{fig:datasetSplit}.

\begin{table*}
    \centering
    \caption{The statistical information of the experimental datasets.}
    \label{table:datasetSplit}
    \resizebox{\textwidth}{!}{
    \begin{tabular}{c|ccccccc}
    \hline
        \textbf{dataset} & \textbf{nodes} & \textbf{features} & \textbf{edges} & \textbf{classes} & \textbf{train/val/test} & \textbf{description} \\ \hline
        Cora & 2708 & 1433 & 5429 & 7 & 20\%/40\%/40\% & citation network \\ 
        CiteSeer & 3327 & 3703 & 4732 & 6 & 20\%/40\%/40\% & citation network \\ 
        PubMed & 19717 & 500 & 44338 & 3 & 20\%/40\%/40\% & citation network \\ \hline
        Amazon Photo & 7487 & 745 & 119043 & 8 & 20\%/40\%/40\% & co-purchase graph \\ 
        Amazon Computer & 13381 & 767 & 245778 & 10 & 20\%/40\%/40\% & co-purchase graph \\ \hline
        Coauthor CS & 18333 & 6805 & 81894 & 15 & 20\%/40\%/40\% & co-authorship graph \\ 
        Coauthor Physics & 34493 & 8415 & 247962 & 5 & 20\%/40\%/40\% & co-authorship graph \\ \hline
        ogbn-arxiv & 169343 & 128 & 1166243 & 40 & 60\%/20\%/20\% & citation network \\ \hline
    \end{tabular}}
\end{table*}

\textbf{Baselines.} We compare our method against ten baselines. For traditional FL, we select FedAvg~\cite{mcmahan2017communication}, FedProx~\cite{li2020federated}, and MOON~\cite{li2021model}. For FGL, we include FedSage+~\cite{zhang2021subgraph}, FedGTA~\cite{li2024FedGTA}, and FedTAD~\cite{zhu2024fedtad}. For AFL, we compare with FedAsyn~\cite{xie2019asynchronous}, TWAFL~\cite{chen2019communication}, FedBuff~\cite{nguyen2022federated}, and SWESALT~\cite{liao2023accelerating}. All experiments are conducted using the GCN model as the backbone architecture.

\textbf{Implementation \& Hyperparameters.} Our training scripts are developed using the open-source FGL framework OpenFGL\cite{li2024openfgl}. Baseline structural hyperparameters follow their original papers. Across all methods, local model optimization uniformly employs the Adam optimizer with a learning rate of 0.001, a weight decay of $5 \times 10^{-4}$, and a batch size of 128. Since SWESALT targets self-supervised learning, we adapt its local objective to a standard cross-entropy loss to ensure fair comparison in our supervised node classification tasks. For FedSA-GCL, the semantic clustering threshold $\theta$ in Eq.~\eqref{eq:sim} is set to 0.5, and the staleness attenuation coefficient $\alpha$ in Eq.~\eqref{eq:omega_server} is configured to 0.4. Additionally, for the Non-param LP steps in Eq.~\eqref{eq:LP}, the propagation balance parameter $\lambda$ is empirically set to 0.5, and the number of propagation steps is fixed at $k = 5$. Model validation is performed locally on each client's dataset.

\textbf{Asynchronous Simulation.} Because synchronous and asynchronous approaches differ in their definition of a training round, we adopt the client trip\cite{nguyen2022federated} as a unified evaluation metric. A client trip is defined as a complete cycle that culminates in a upstream transmission of local model updates. This cycle includes retrieving the global model, executing a single epoch of local training, and uploading the computed parameters. We evaluate all models over 2000 client trips. We simulate the asynchronous environment at the system level using the Python \texttt{threading} library, treating each client as an independent thread to mimic concurrent real-world arrivals. The server maintains a thread-safe message queue and triggers a global semi-asynchronous aggregation whenever it accumulates $K=5$ client updates. To emulate hardware and network heterogeneity within this multi-threaded implementation, 30\% of the clients are randomly designated as stragglers. By programmatically suspending their respective threads, these edge devices are delayed by 2 to 5 complete global training cycles, where one global cycle comprises a number of client trips equal to the total number of participating clients.

\textbf{Experiment Environment.} All experiments are conducted on a machine equipped with an Intel Core i9-13900K processor @5.8GHz and an NVIDIA GeForce RTX 3090 GPU with 24GB memory, running CUDA 12.1. The operating system is Ubuntu 22.04, and the system has 64GB of RAM.

\begin{table*}[ht]
    \centering
    \caption{Performance comparison under Louvain splitting.}
    \label{table: Performance louvain}
    \resizebox{\textwidth}{!}{
    \begin{tabular}{c|c|ccccccccc}
    \hline
        \textbf{type}  &  \textbf{Method}  &  \textbf{Cora}  &  \textbf{CiteSeer}  &  \textbf{PubMed}  &  \textbf{CS}  &  \textbf{Photo}  &  \textbf{Computers}  &  \textbf{Physics}  &  \textbf{ogbn-arxiv}  & \textbf{Overall} \\ \hline
        ~  &  FedAvg  & $69.0_{\pm 1.3}$ & $71.8_{\pm 0.5}$ & $82.9_{\pm 0.6}$ & $79.1_{\pm 3.2}$ & $67.2_{\pm 5.3}$ & $62.1_{\pm 7.2}$ & $88.0_{\pm 1.3}$ & $52.6_{\pm 0.6}$ & $71.6_{\pm 2.5}$ \\
        FL  &  FedProx  & $67.6_{\pm 1.4}$ & $71.2_{\pm 0.3}$ & $82.8_{\pm 0.7}$ & $78.3_{\pm 1.5}$ & $64.1_{\pm 3.8}$ & $62.9_{\pm 2.0}$ & $88.9_{\pm 2.1}$ & $51.6_{\pm 0.8}$ & $70.9_{\pm 1.6}$ \\
        ~  &  MOON  & $56.4_{\pm 1.3}$ & $69.2_{\pm 5.8}$ & $65.6_{\pm 1.2}$ & $70.4_{\pm 1.7}$ & $66.5_{\pm 5.4}$ & $64.0_{\pm 3.5}$ & $85.8_{\pm 1.2}$ & $49.9_{\pm 1.5}$ & $66.0_{\pm 2.7}$ \\ \hline
        ~  &  FedSage+  & $68.3_{\pm 1.7}$ & $69.6_{\pm 0.6}$ & $81.7_{\pm 4.3}$ & $78.7_{\pm 3.2}$ & $72.9_{\pm 6.2}$ & $65.6_{\pm 3.8}$ & $88.5_{\pm 0.5}$ & $53.2_{\pm 1.4}$ & $72.3_{\pm 2.7}$ \\
        FGL  &  FedGTA  & $74.0_{\pm 2.7}$ & $71.4_{\pm 0.5}$ & $82.6_{\pm 0.4}$ & $\mathbf{88.4}_{\pm 0.2}$ & $82.4_{\pm 0.6}$ & $73.9_{\pm 0.6}$ & $\mathbf{93.7}_{\pm 0.2}$ & $58.7_{\pm 0.5}$ & $78.1_{\pm 0.7}$ \\
        ~  &  FedTAD  & $63.6_{\pm 9.8}$ & $70.9_{\pm 0.4}$ & $77.2_{\pm 8.2}$ & $71.0_{\pm 1.3}$ & $65.8_{\pm 5.5}$ & $62.7_{\pm 2.9}$ & $87.0_{\pm 0.3}$ & $50.4_{\pm 2.2}$ & $68.6_{\pm 3.8}$ \\ \hline
        ~  &  FedAsyn  & $73.4_{\pm 0.7}$ & $70.9_{\pm 0.4}$ & $71.1_{\pm 0.4}$ & $74.7_{\pm 1.5}$ & $77.5_{\pm 0.9}$ & $70.1_{\pm 0.7}$ & $88.3_{\pm 0.4}$ & $40.8_{\pm 0.5}$ & $70.8_{\pm 0.7}$ \\
        AFL  &  TWAFL  & $69.3_{\pm 0.7}$ & $47.8_{\pm 1.0}$ & $64.1_{\pm 1.2}$ & $61.5_{\pm 3.6}$ & $74.0_{\pm 1.8}$ & $66.0_{\pm 2.1}$ & $77.5_{\pm 1.7}$ & $29.6_{\pm 1.1}$ & $61.2_{\pm 1.6}$ \\
        ~  &  FedBuff  & $73.8_{\pm 1.0}$ & $70.6_{\pm 0.3}$ & $84.1_{\pm 0.1}$ & $81.2_{\pm 0.9}$ & $73.3_{\pm 4.4}$ & $69.3_{\pm 2.2}$ & $92.0_{\pm 1.4}$ & $52.5_{\pm 0.6}$ & $74.6_{\pm 1.4}$ \\
        ~  &  SWESALT  & $73.9_{\pm 1.6}$ & $\mathbf{73.1}_{\pm 0.3}$ & $82.7_{\pm 0.3}$ & $81.0_{\pm 2.5}$ & $77.9_{\pm 1.3}$ & $68.4_{\pm 0.4}$ & $90.1_{\pm 2.4}$ & $44.3_{\pm 1.9}$ & $73.9_{\pm 1.3}$ \\ \hline
        ~  &  FedSA-GCL  & $\mathbf{75.8}_{\pm 1.6}$ & $71.8_{\pm 0.5}$ & $\mathbf{85.0}_{\pm 0.2}$ & $85.8_{\pm 0.1}$ & $\mathbf{87.6}_{\pm 1.5}$ & $\mathbf{82.6}_{\pm 1.0}$ & $92.9_{\pm 0.5}$ & $\mathbf{58.7}_{\pm 0.3}$ & $\mathbf{80.0}_{\pm 2.3}$ \\ \hline
    \end{tabular}}
\end{table*}

\begin{table*}[ht]
    \centering
    \caption{Performance comparison under Metis splitting.}
    \label{table: Performance metis}
    \resizebox{\textwidth}{!}{
    \begin{tabular}{c|c|ccccccccc}
    \hline
        \textbf{type}  &  \textbf{Method}  &  \textbf{Cora}  &  \textbf{CiteSeer}  &  \textbf{PubMed}  &  \textbf{CS}  &  \textbf{Photo}  &  \textbf{Computers}  &  \textbf{Physics}  &  \textbf{ogbn-arxiv}  & \textbf{Overall} \\ \hline
        ~  &  FedAvg  & $63.0_{\pm 2.7}$ & $71.6_{\pm 0.9}$ & $81.8_{\pm 0.4}$ & $71.0_{\pm 3.0}$ & $74.3_{\pm 6.9}$ & $56.6_{\pm 5.2}$ & $87.9_{\pm 0.3}$ & $54.4_{\pm 0.2}$ & $70.1_{\pm 2.5}$ \\
        FL  &  FedProx  & $56.7_{\pm 4.4}$ & $71.5_{\pm 0.8}$ & $82.3_{\pm 0.6}$ & $70.3_{\pm 2.8}$ & $73.7_{\pm 5.3}$ & $57.6_{\pm 6.2}$ & $87.8_{\pm 0.4}$ & $54.5_{\pm 1.2}$ & $69.3_{\pm 2.7}$ \\
        ~  &  MOON  & $70.4_{\pm 4.3}$ & $71.1_{\pm 0.5}$ & $77.9_{\pm 6.5}$ & $85.0_{\pm 2.2}$ & $74.8_{\pm 0.8}$ & $66.1_{\pm 2.5}$ & $87.8_{\pm 0.5}$ & $53.8_{\pm 1.8}$ & $73.4_{\pm 2.4}$ \\ \hline
        ~  &  FedSage+  & $73.7_{\pm 0.8}$ & $69.6_{\pm 0.7}$ & $82.5_{\pm 1.8}$ & $78.5_{\pm 0.9}$ & $80.9_{\pm 0.2}$ & $71.6_{\pm 1.7}$ & $89.4_{\pm 1.3}$ & $57.4_{\pm 1.0}$ & $75.5_{\pm 1.0}$ \\
        FGL  &  FedGTA  & $73.2_{\pm 2.4}$ & $71.4_{\pm 0.7}$ & $79.5_{\pm 2.3}$ & $\mathbf{89.3}_{\pm 0.2}$ & $82.9_{\pm 1.3}$ & $73.2_{\pm 0.8}$ & $94.0_{\pm 0.3}$ & $59.9_{\pm 0.2}$ & $77.9_{\pm 1.0}$ \\
        ~  &  FedTAD  & $50.3_{\pm 8.5}$ & $71.2_{\pm 0.7}$ & $70.2_{\pm 6.5}$ & $67.7_{\pm 3.2}$ & $70.2_{\pm 5.9}$ & $63.6_{\pm 3.8}$ & $86.9_{\pm 1.0}$ & $52.1_{\pm 2.5}$ & $66.5_{\pm 4.0}$ \\ \hline
        ~  &  FedAsyn  & $77.7_{\pm 0.3}$ & $71.2_{\pm 0.2}$ & $76.6_{\pm 1.1}$ & $79.1_{\pm 1.5}$ & $81.8_{\pm 0.7}$ & $73.9_{\pm 1.2}$ & $88.3_{\pm 0.4}$ & $48.1_{\pm 0.6}$ & $74.6_{\pm 0.8}$ \\
        AFL  &  TWAFL  & $65.7_{\pm 1.0}$ & $48.1_{\pm 2.3}$ & $71.4_{\pm 0.4}$ & $70.1_{\pm 1.2}$ & $78.4_{\pm 1.4}$ & $67.3_{\pm 0.8}$ & $83.5_{\pm 0.7}$ & $42.8_{\pm 0.9}$ & $65.9_{\pm 1.1}$ \\
        ~  &  FedBuff  & $75.1_{\pm 0.4}$ & $70.4_{\pm 0.4}$ & $\mathbf{83.9}_{\pm 0.3}$ & $85.9_{\pm 0.4}$ & $76.5_{\pm 3.6}$ & $66.2_{\pm 1.3}$ & $88.2_{\pm 0.3}$ & $54.7_{\pm 0.4}$ & $75.1_{\pm 0.9}$ \\
        ~  &  SWESALT  & $73.6_{\pm 1.2}$ & $70.1_{\pm 0.6}$ & $81.1_{\pm 0.3}$ & $77.6_{\pm 0.2}$ & $80.3_{\pm 0.3}$ & $72.7_{\pm 1.3}$ & $87.7_{\pm 0.6}$ & $43.0_{\pm 3.1}$ & $73.3_{\pm 1.0}$ \\ \hline
        ~  &  FedSA-GCL  & $\mathbf{81.7}_{\pm 0.3}$ & $\mathbf{71.7}_{\pm 1.7}$ & $83.6_{\pm 0.3}$ & $89.2_{\pm 0.3}$ & $\mathbf{85.5}_{\pm 0.9}$ & $\mathbf{80.3}_{\pm 4.2}$ & $\mathbf{94.0}_{\pm 0.1}$ & $\mathbf{61.5}_{\pm 0.3}$ & $\mathbf{80.9}_{\pm 1.0}$ \\ \hline
    \end{tabular}}
\end{table*}

\subsection{Performance Comparison}

To answer \textbf{Q1}, we conduct experiments on various datasets split into 20 clients using different community detection algorithms. Each baseline is trained five times, and we report the mean performance along with the 95\% confidence interval. Our method outperforms the baselines in the vast majority of cases. Specifically, compared to the strongest baseline, our method achieves an absolute average improvement of 1.9\% under the Louvain splitting, and 3.0\% under the Metis splitting. 
Detailed results are presented in Table~\ref{table: Performance louvain} and Table~\ref{table: Performance metis}, where the best results are highlighted using bold.

\begin{figure}[ht]
    
    \centering
    \includegraphics[width=1\linewidth]{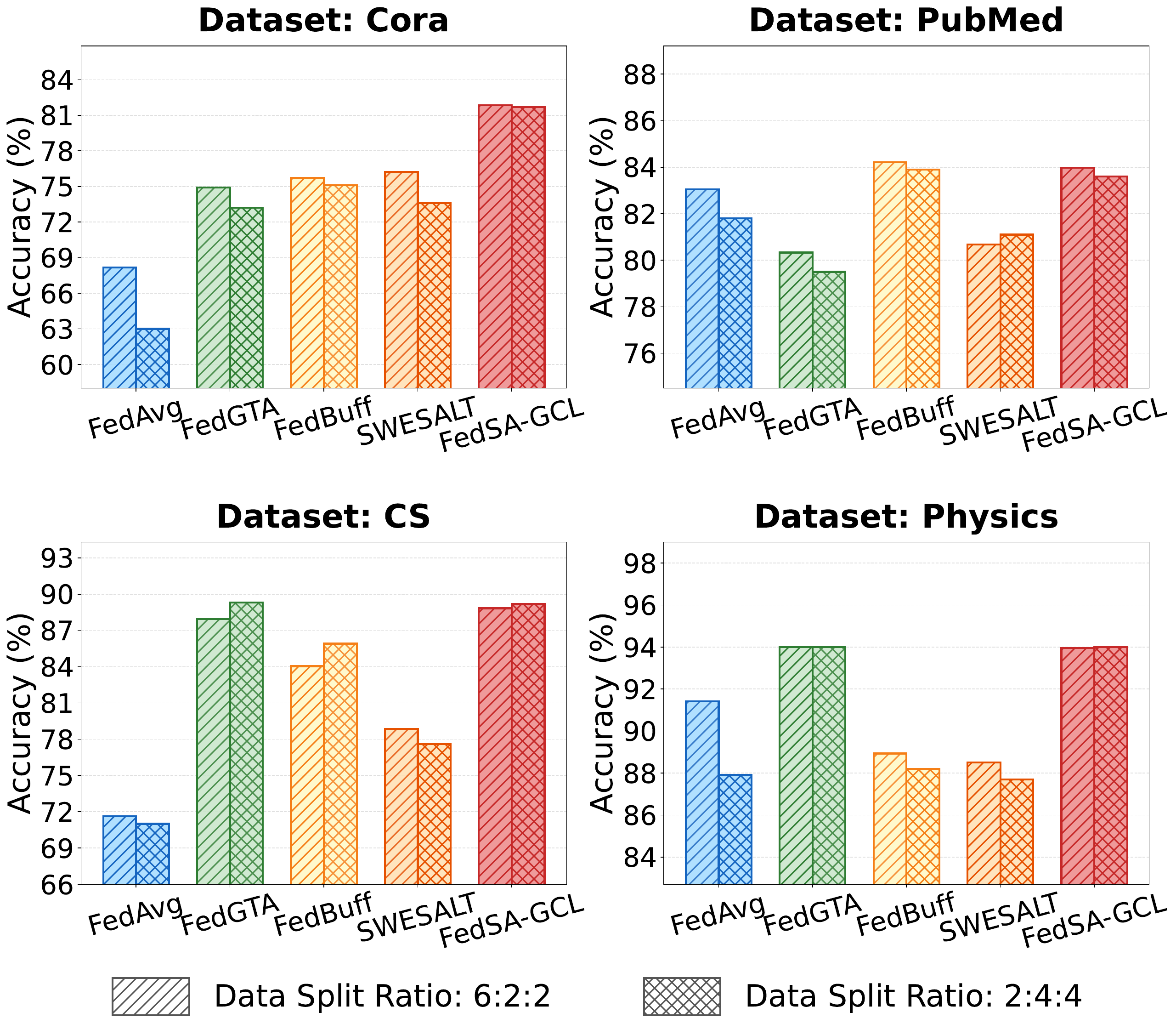}
    \caption{Performance comparison of different methods under 2:4:4 and 6:2:2 data splits.}
    \label{fig:split_ratio}
    
\end{figure}

Due to the inherent structural dependencies in graph data, a modest fraction of labeled data typically suffices for model fitting, making the 2:4:4 partition a standard configuration in prior research \cite{kipf2016semi,shchur2018pitfalls}. To further validate this characteristic, we compare the 2:4:4 and 6:2:2 splits across multiple baseline methods on Louvain-partitioned subgraphs, as illustrated in Fig.~\ref{fig:split_ratio}. While the 6:2:2 split generally yields higher accuracy, the overall performance gap remains negligible. Notably, exposing models to larger local training sets can sometimes induce topological overfitting, a phenomenon evidenced by the slight performance drops of SWESALT on the PubMed dataset and FedBuff on the CS dataset under the 6:2:2 configuration. Conversely, FedAvg suffers substantial performance degradation when the training data ratio is reduced, particularly on the Cora dataset, primarily because it lacks dedicated strategies to mitigate the severe non-IID challenges inherent in federated graphs.

\begin{figure*}[htbp]
    \centering
    \subfigure[Cross-Method Comparison of Client Similarity Evolution on Cora\label{fig:Similarity Cora}]{\includegraphics[width=0.95\textwidth]{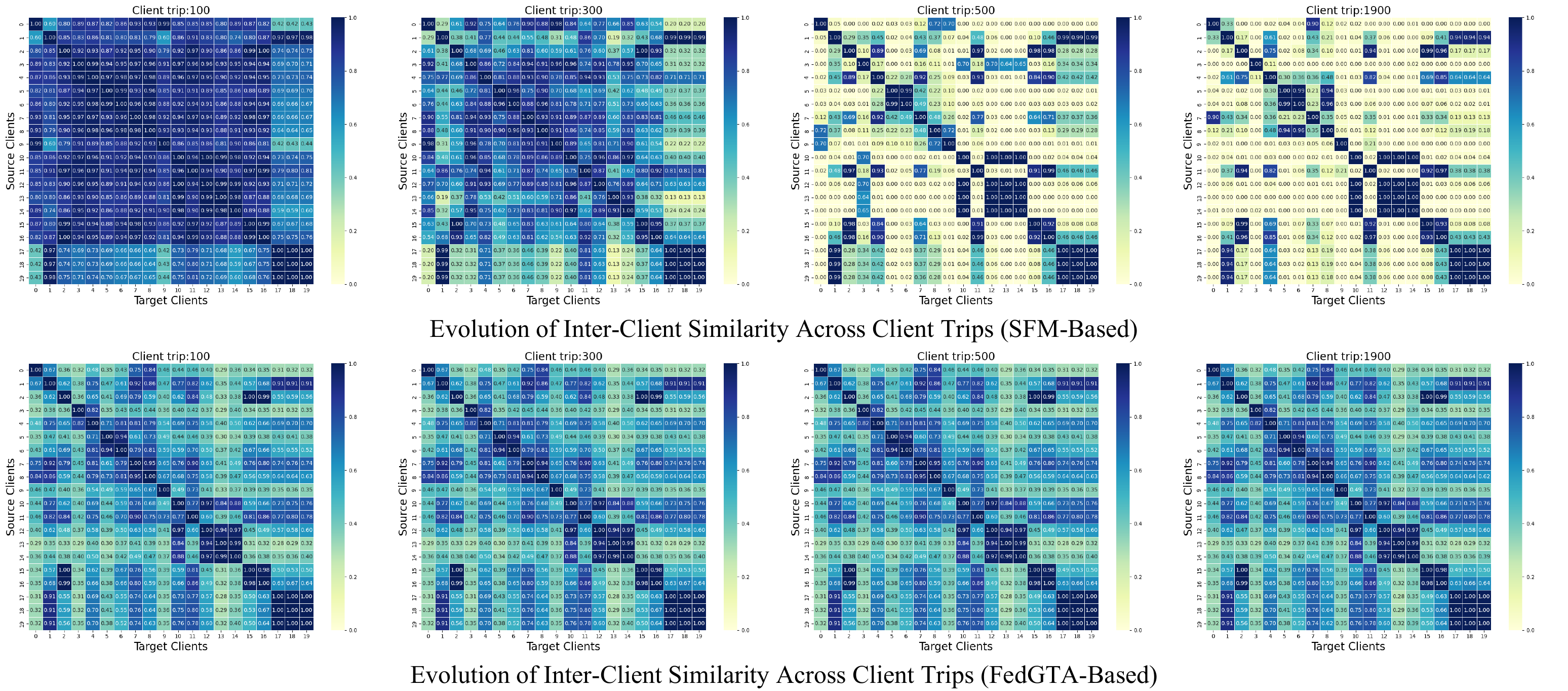} }
    \subfigure[Cross-Method Comparison of Client Similarity Evolution on PubMed\label{fig:Similarity PubMed}]{\includegraphics[width=0.95\textwidth]{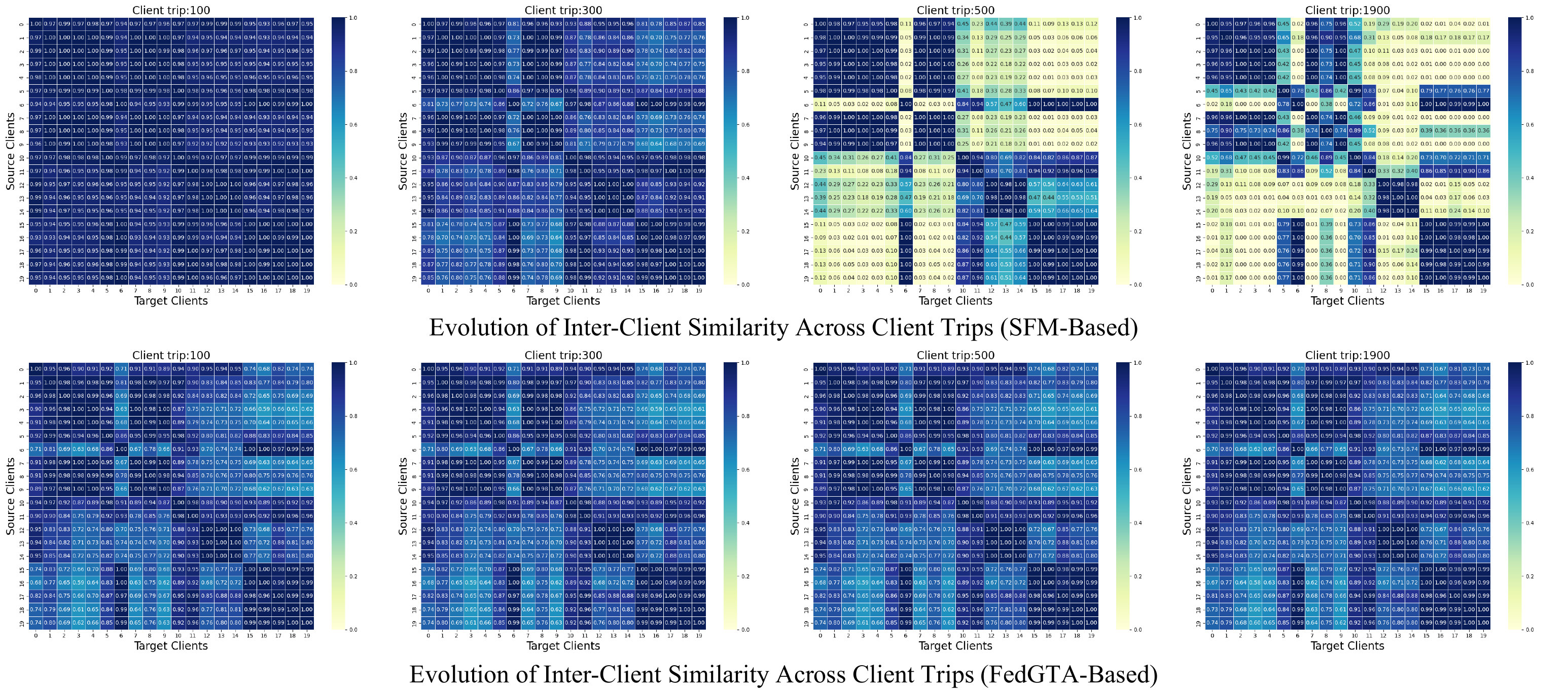} }
    \caption{Client Similarity Evolution Across Client Trips on Cora and PubMed.}
    \label{fig:ablation study2}
\end{figure*}

\subsection{Method Interpretability}
\label{sec:Interpretability}

To answer \textbf{Q2}, we examine the implementation details of our method and conduct ablation studies on two representative datasets, Cora and PubMed, both of which are split using the Metis community detection algorithm. The objective of these experiments is to validate the following points:
\textbf{Q2-1.} The LSC aggregation weights, which incorporate staleness-aware penalties, can accelerate convergence in asynchronous federated graph learning.
\textbf{Q2-2.} The proposed ClusterCast mechanism is well-suited for asynchronous federated graph scenarios and improves overall accuracy.
\textbf{Q2-3.} Our proposed clustering algorithm based on the SFM can effectively group clients and improve overall convergence speed.

\begin{figure}
    \centering
    \subfigure[Training Curve under Different Ablation Settings on Cora.\label{fig:Training Curve Cora}]{
        \includegraphics[width=0.45\textwidth]{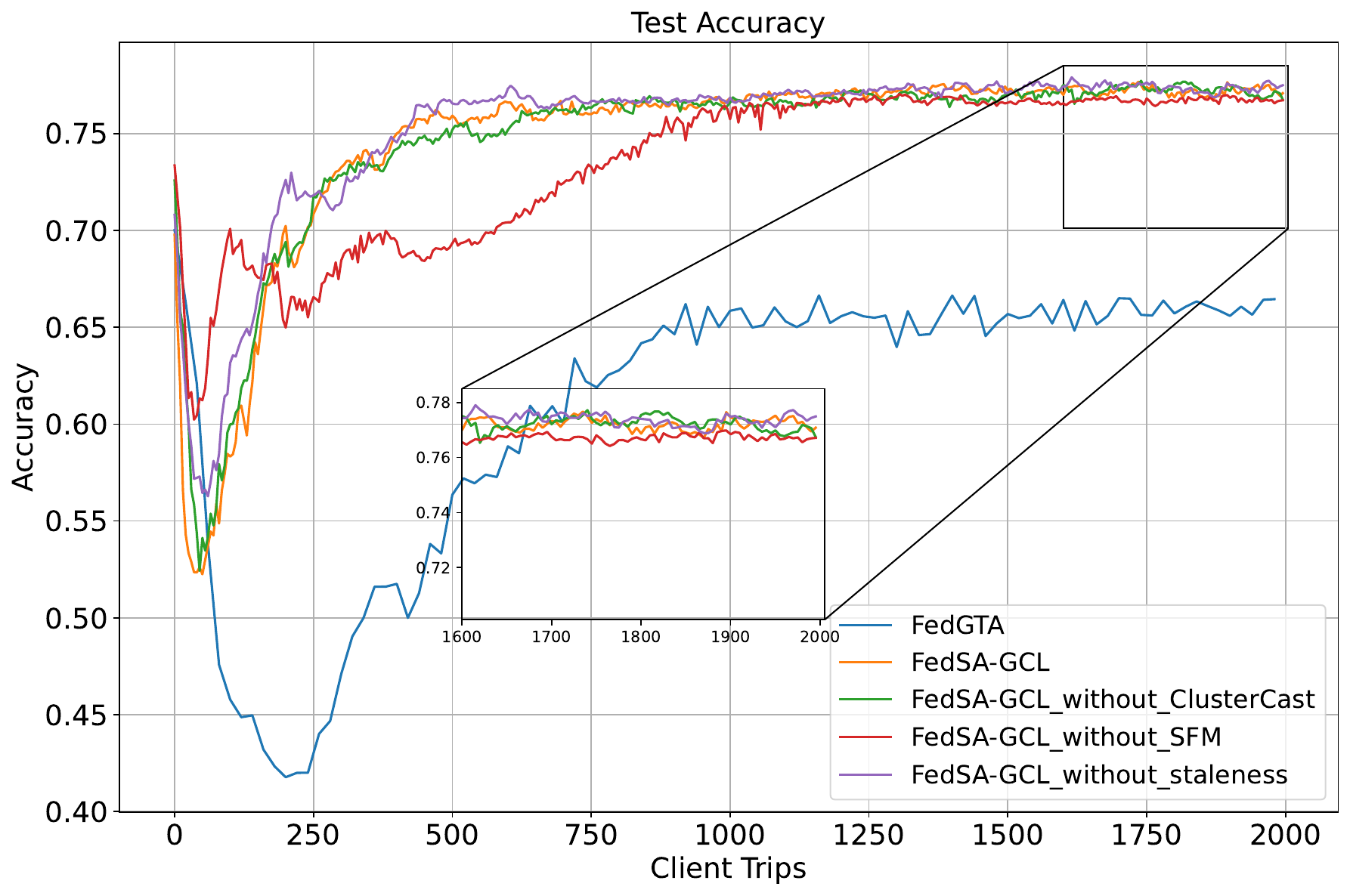}
    }
    \subfigure[Training Curve under Different Ablation Settings on PubMed.\label{fig:Training Curve PubMed}]{
        \includegraphics[width=0.45\textwidth]{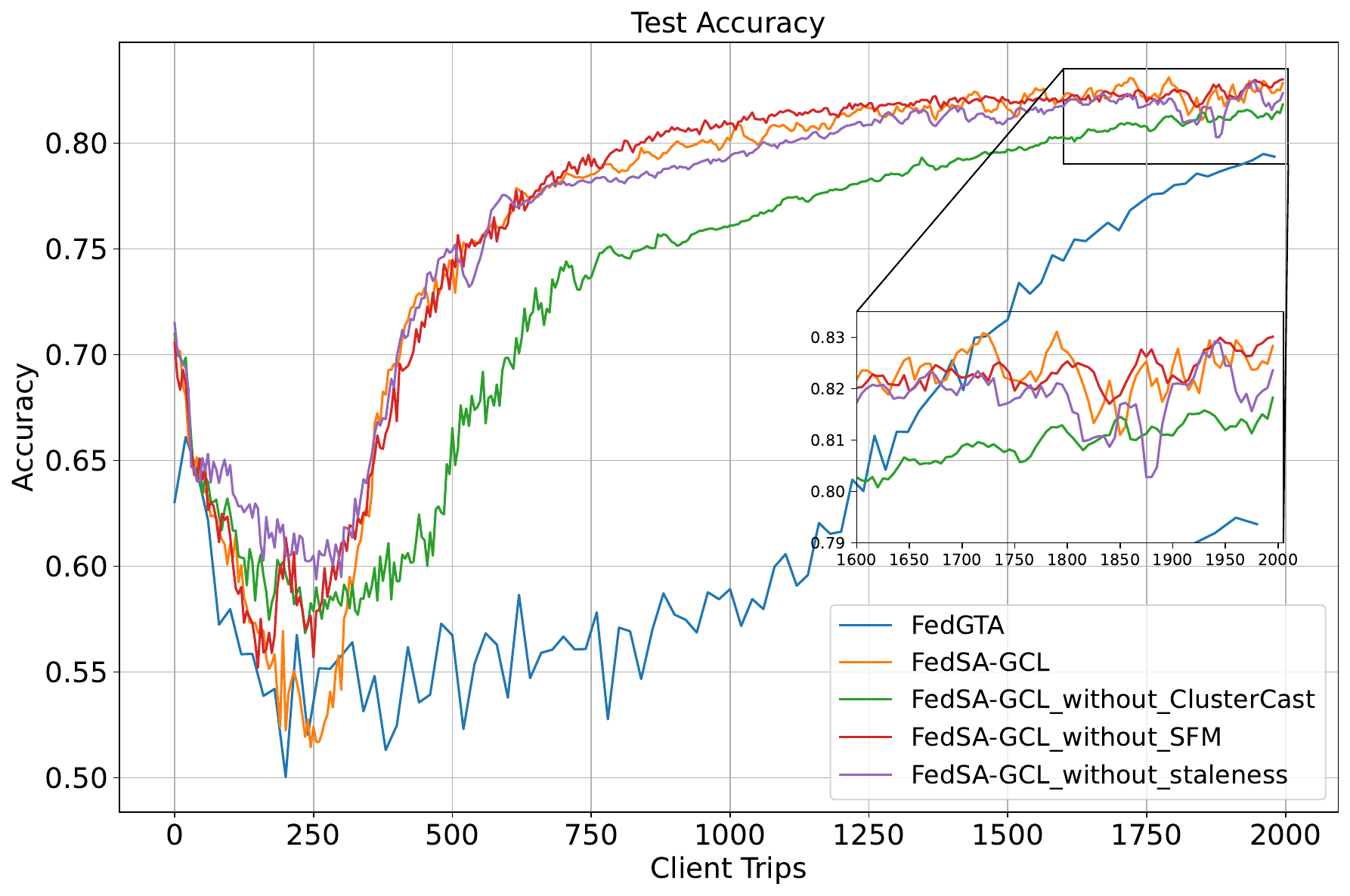}
    }
    
    \caption{Ablation Performance Comparison on Cora and PubMed.}
    \label{fig:ablation study}
\end{figure}

Regarding \textbf{Q2-1}, this phenomenon is demonstrated by the orange and purple curves in Fig.~\ref{fig:Training Curve Cora} and Fig.~\ref{fig:Training Curve PubMed}. Without the incorporation of staleness-aware LSC aggregation weights, the influence of stale models leads to unstable convergence in the later training stages, thereby slowing down the overall convergence process.

For \textbf{Q2-2}, the advantage of the ClusterCast mechanism is illustrated by comparing the orange and green curves in Fig.~\ref{fig:Training Curve Cora} and Fig.~\ref{fig:Training Curve PubMed}. The results clearly show that incorporating ClusterCast significantly enhances both the convergence speed and final accuracy of the model.

As for \textbf{Q2-3}, this can be observed through the orange and red curves in Fig.~\ref{fig:Training Curve Cora} and Fig.~\ref{fig:Training Curve PubMed}, as well as from Fig.~\ref{fig:Similarity Cora} and Fig.~\ref{fig:Similarity PubMed}. By comparing client similarity heatmaps across different client trips, it can be seen that the clustering behavior of FedGTA remains largely consistent across client trips, while our SFM-based clustering method responds more dynamically. Notably, in the PubMed dataset, FedGTA fails to form effective client clusters. Although FedGTA's clustering appears more confident in the early training stages and results in faster initial convergence due to more accurate client similarity estimation, our method gradually improves its clustering confidence over time. As a result, model accuracy stabilizes and ultimately converges. This trend is particularly evident in Fig.~\ref{fig:Training Curve Cora}, which shows that our clustering method achieves slightly higher accuracy on Cora and comparable accuracy on PubMed.

\subsection{Hyperparameter Sensitivity}
\label{sec:Hyperparameter Sensitivity}

To answer \textbf{Q3}, we investigate the sensitivity of the proposed framework to critical hyperparameters. We use Louvain to split client subgraphs. Specifically, we first analyze the trade-off between local epochs and communication rounds to understand its impact on training dynamics and convergence efficiency. Subsequently, we evaluate the stability of the model under varying values of $\alpha$ and the similarity threshold $\theta$. 

\begin{figure}[H]
    \centering
    \subfigure[Cora Dateset Test Accuracy under Different Local Epochs]{
        \includegraphics[width=0.45\textwidth]{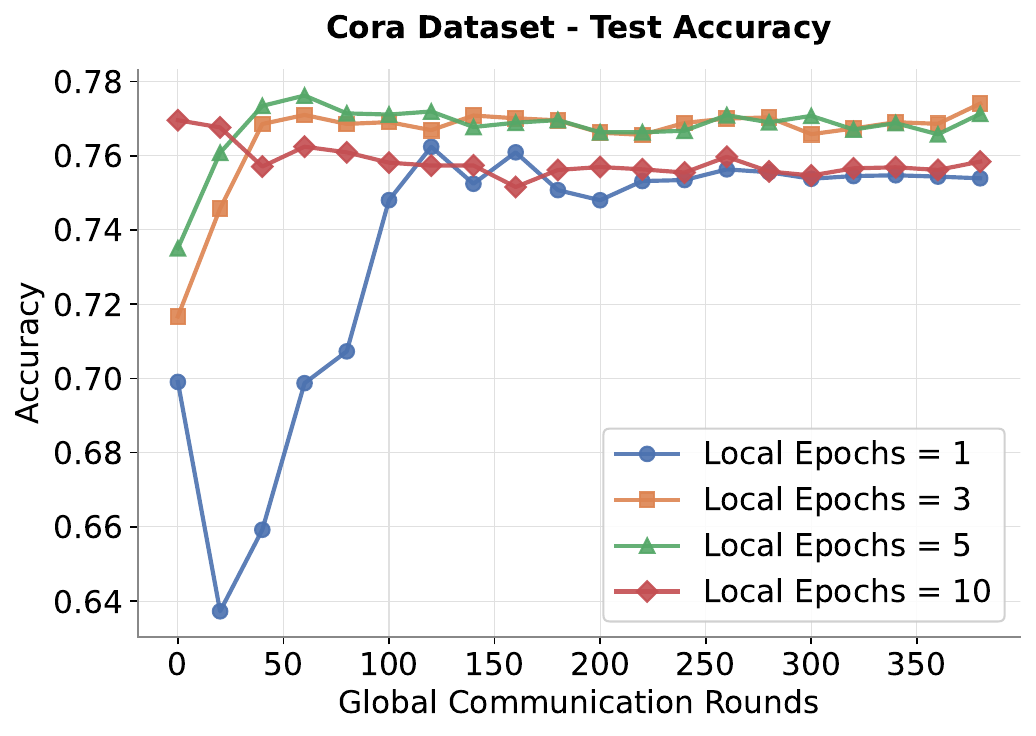}
        \label{fig:cora_epoch}
    }
    \subfigure[Computers Dataset Test Accuracy under Different Local Epochs]{
        \includegraphics[width=0.45\textwidth]{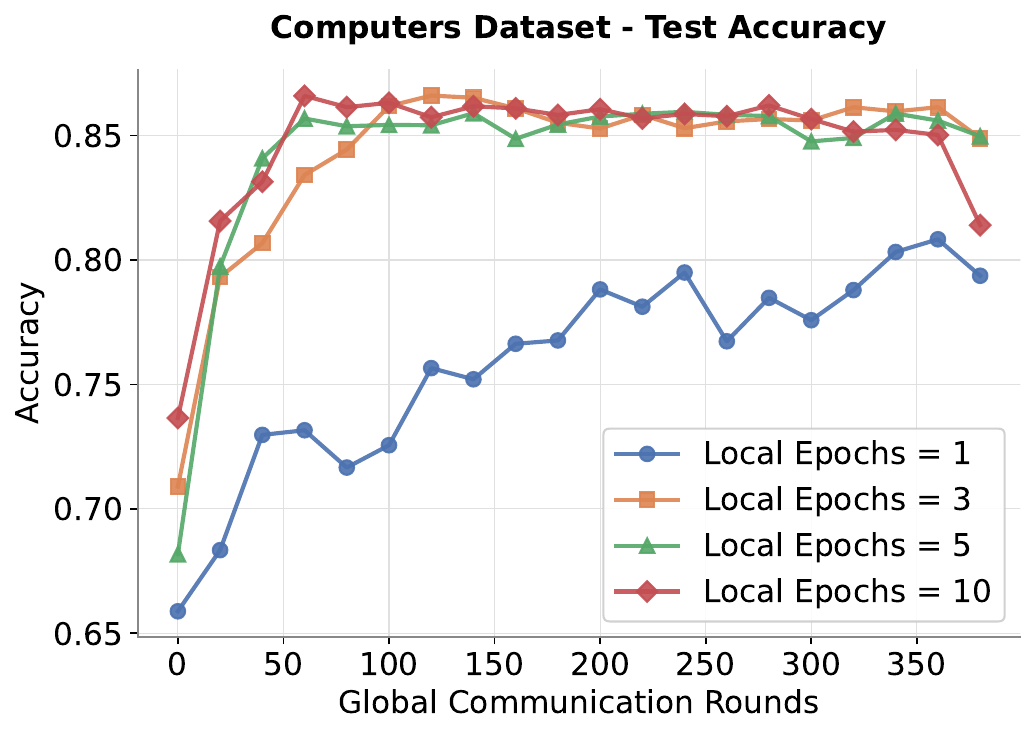}
        \label{fig:computers_epoch}
    }
    \caption{Impact of local epochs on model performance across different datasets.}
    \label{fig:epoch_analysis}
\end{figure}

\textbf{Local epochs vs. communication rounds.} 
The impact of the number of local epochs on model performance is illustrated in Fig.~\ref{fig:epoch_analysis}. On the Cora dataset (Fig.~\ref{fig:cora_epoch}), setting the local epoch to 1 results in slower initial convergence and slightly lower final accuracy. However, the performance gap compared to 3, 5, and 10 epochs remains marginal. When the local epoch is increased to 10, the model converges rapidly in the early stages but subsequently experiences a minor decline in accuracy. Consequently, the overall performance of 10 epochs is lower than that of 3 and 5 epochs.

The Computers dataset (Fig.~\ref{fig:computers_epoch}) features a larger graph scale with denser edges. On this dataset, a local epoch of 1 leads to relatively lower accuracy, yet the model still achieves convergence with a final accuracy of approximately 80\%. Notably, setting the epoch to 10 results in a sharp decrease in accuracy toward the end of training. Similar to the observations on the Cora dataset, this trend indicates potential overfitting due to excessive local training steps.

\begin{figure*}
    \centering
    \subfigure[$\alpha$ Sensitivity: Accuracy vs. Communication Cost Trade-off.]{
        \includegraphics[width=1\textwidth]{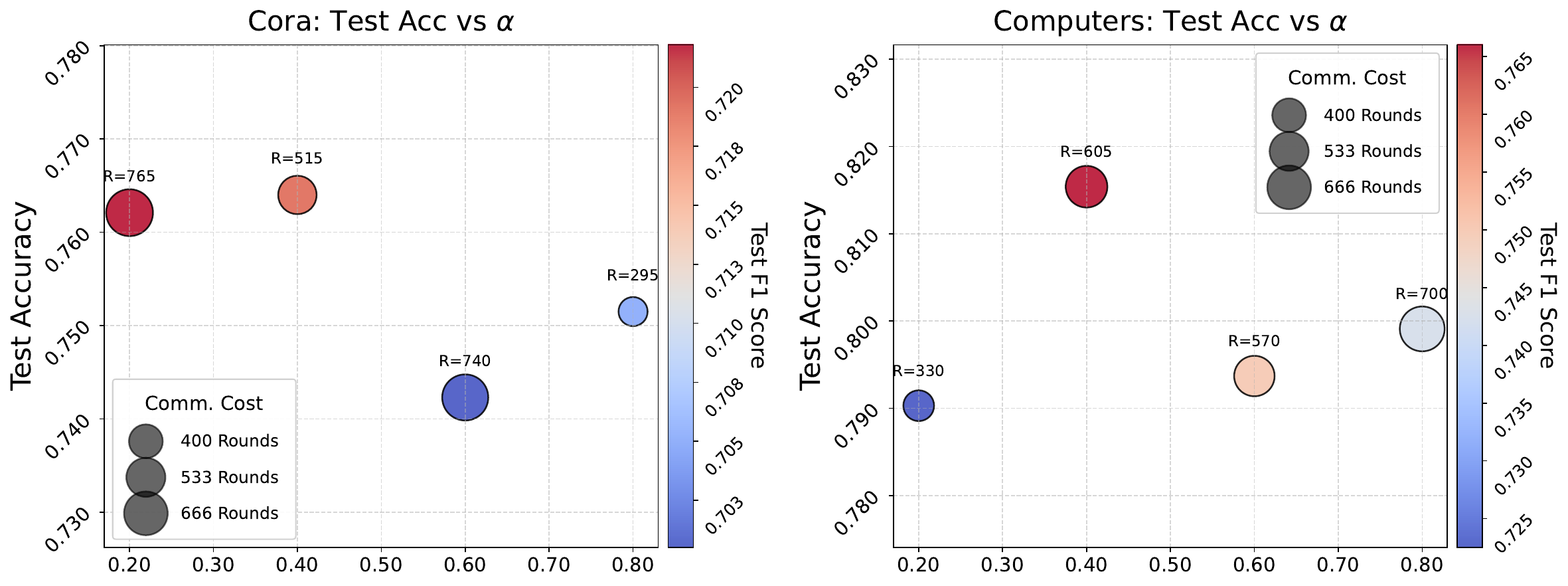}
        \label{fig:bubble_alpha}
    }
    \subfigure[$\theta$ Sensitivity: Accuracy vs. Communication Cost Trade-off.]{
        \includegraphics[width=1\textwidth]{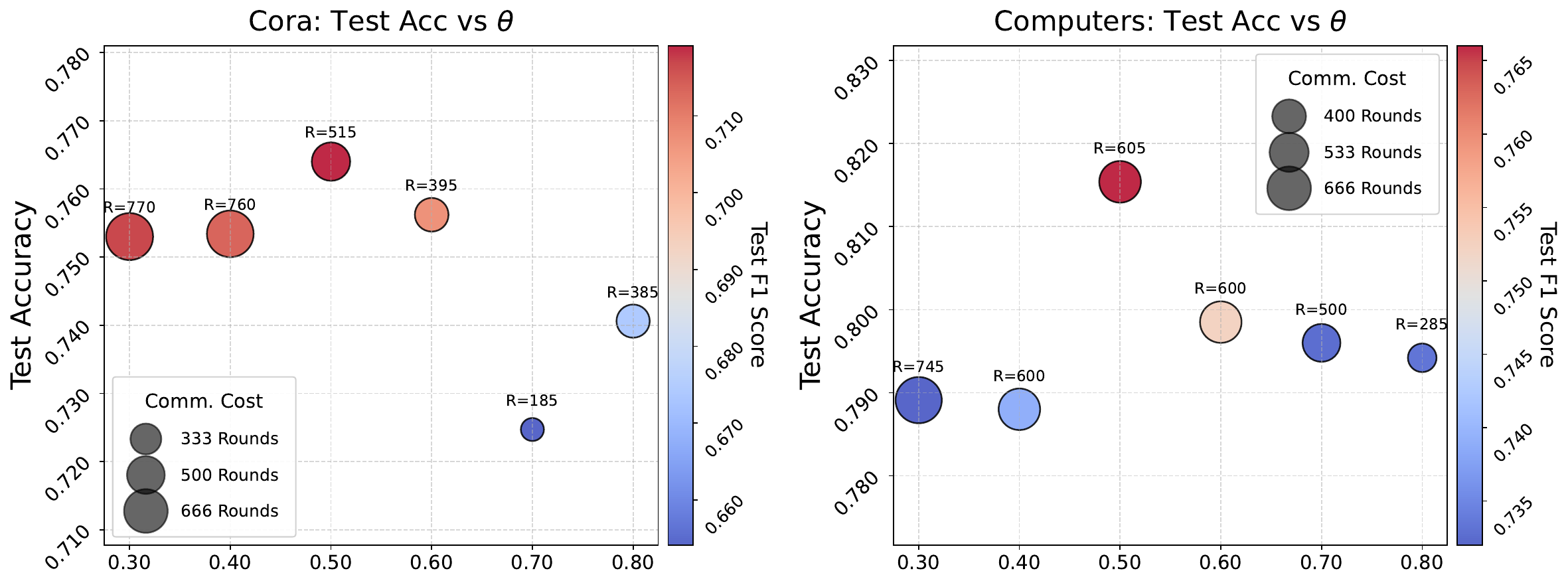}
        \label{fig:bubble_theta}
    }
    \caption{Sensitivity analysis of hyperparameters $\alpha$ and $\theta$.}
    \label{fig:hyper_sensitivity}
\end{figure*}

\textbf{Sensitivity of $\alpha$ and Similarity threshold ($\theta$).} 
Fig.~\ref{fig:hyper_sensitivity} presents the sensitivity analysis for hyperparameters $\alpha$ and $\theta$. Under our predefined parameter settings, the model achieves maximum accuracy while maintaining moderate communication costs, indicating stable convergence behavior. Regarding the similarity threshold $\theta$ in Eq.~\eqref{eq:sim} (Fig.~\ref{fig:bubble_theta}), setting the value to 0.5 is theoretically grounded. Since the soft label vectors are non-negative, their cosine similarity strictly ranges from 0 to 1, making 0.5 the unbiased mathematical midpoint. Our empirical results confirm that this static value yields optimal performance. Furthermore, employing an adaptive $\theta$ strategy would necessitate continuous global state synchronization, thereby introducing severe communication bottlenecks that undermine the efficiency of the asynchronous design.

\subsection{Robustness Comparison}

To answer \textbf{Q4}, we conduct robustness experiments on eight graph datasets using Metis to split client subgraphs. We introduce two types of perturbations to evaluate algorithm stability: random label sparsity, achieved by uniformly masking 10\% of local training labels (Table~\ref{table:label sparsity}), and random topology sparsity, implemented by symmetrically dropping 10\% of edges (Table~\ref{table:topology sparsity}). To ensure fair comparison, these identically perturbed datasets are consistently applied across all baselines. Each baseline is trained five times. We report the mean performance along with the 95\% confidence interval.

Under the random topology sparsity setting, our method outperforms the best-performing baseline by an absolute average of 2.8\%. In the random label sparsity scenario, the absolute improvement reaches 3.7\%. Specifically, we observe the model performance on the large-scale \texttt{ogbn-arxiv} dataset. FedSA-GCL maintains the highest test accuracy under both perturbation types. It demonstrates stronger resilience against data noise compared to the asynchronous baseline SWESALT. These results confirm the robustness of our proposed method against structural and annotation perturbations.

\begin{table*}
    \centering
    \caption{Performance under Random Label Sparsity}
    \label{table:label sparsity}
    \resizebox{\textwidth}{!}{
    \begin{tabular}{c|c|ccccccccc}
    \hline
        \textbf{type} & \textbf{Method} & \textbf{Cora} & \textbf{CiteSeer} & \textbf{PubMed} & \textbf{CS} & \textbf{Photo} & \textbf{Computers} & \textbf{Physics} & \textbf{ogbn-arxiv} & \textbf{Overall} \\ \hline
        ~ & FedAvg & $63.6_{\pm 2.2}$ & $70.6_{\pm 0.4}$ & $81.2_{\pm 1.3}$ & $72.4_{\pm 4.8}$ & $79.6_{\pm 3.0}$ & $59.9_{\pm 3.5}$ & $87.5_{\pm 0.0}$ & $53.9_{\pm 0.4}$ & $71.1_{\pm 3.3}$ \\ 
        FL & FedProx & $61.0_{\pm 3.4}$ & $70.2_{\pm 0.6}$ & $81.9_{\pm 0.5}$ & $74.0_{\pm 3.7}$ & $76.6_{\pm 6.3}$ & $55.3_{\pm 2.9}$ & $87.9_{\pm 0.6}$ & $54.6_{\pm 0.9}$ & $70.2_{\pm 3.8}$ \\ 
        ~ & MOON & $64.0_{\pm 5.9}$ & $64.5_{\pm 7.0}$ & $70.6_{\pm 2.1}$ & $65.5_{\pm 3.4}$ & $72.4_{\pm 6.5}$ & $59.4_{\pm 3.8}$ & $87.9_{\pm 0.5}$ & $49.5_{\pm 1.4}$ & $66.7_{\pm 3.2}$ \\ \hline
        ~ & FedSage+ & $73.9_{\pm 1.9}$ & $69.9_{\pm 0.8}$ & $82.1_{\pm 0.8}$ & $78.8_{\pm 1.2}$ & $80.8_{\pm 0.3}$ & $66.8_{\pm 7.0}$ & $89.5_{\pm 2.9}$ & $55.2_{\pm 1.2}$ & $74.6_{\pm 2.6}$ \\ 
        FGL & FedGTA & $71.0_{\pm 1.7}$ & $71.0_{\pm 0.1}$ & $77.5_{\pm 2.1}$ & $\mathbf{89.4}_{\pm 0.3}$ & $82.6_{\pm 0.5}$ & $73.0_{\pm 1.1}$ & $94.0_{\pm 0.2}$ & $60.1_{\pm 0.4}$ & $77.3_{\pm 3.0}$ \\ 
        ~ & FedTAD & $55.1_{\pm 9.2}$ & $\mathbf{72.3}_{\pm 0.5}$ & $70.6_{\pm 6.3}$ & $72.9_{\pm 5.6}$ & $72.9_{\pm 7.6}$ & $63.0_{\pm 0.8}$ & $87.1_{\pm 0.6}$ & $48.5_{\pm 2.1}$ & $67.8_{\pm 3.5}$ \\ \hline
        ~ & FedAsyn & $77.5_{\pm 1.1}$ & $69.4_{\pm 0.4}$ & $74.8_{\pm 1.0}$ & $78.6_{\pm 0.8}$ & $81.9_{\pm 0.7}$ & $74.4_{\pm 1.6}$ & $88.3_{\pm 0.5}$ & $45.7_{\pm 0.6}$ & $73.8_{\pm 2.0}$ \\ 
        AFL & TWAFL & $65.5_{\pm 0.7}$ & $46.0_{\pm 1.1}$ & $71.4_{\pm 0.2}$ & $70.0_{\pm 1.9}$ & $78.8_{\pm 1.4}$ & $67.9_{\pm 3.0}$ & $82.8_{\pm 0.4}$ & $41.7_{\pm 1.0}$ & $65.5_{\pm 3.8}$ \\ 
        ~ & FedBuff & $75.6_{\pm 0.3}$ & $69.4_{\pm 0.2}$ & $83.8_{\pm 0.6}$ & $85.6_{\pm 1.2}$ & $79.3_{\pm 1.4}$ & $67.5_{\pm 1.8}$ & $88.5_{\pm 1.0}$ & $54.1_{\pm 0.2}$ & $75.5_{\pm 2.6}$ \\ 
        ~ & SWESALT & $72.6_{\pm 0.6}$ & $69.1_{\pm 1.2}$ & $80.7_{\pm 0.7}$ & $76.1_{\pm 0.5}$ & $79.3_{\pm 0.3}$ & $73.0_{\pm 1.2}$ & $88.2_{\pm 0.1}$ & $46.0_{\pm 0.5}$ & $73.1_{\pm 0.7}$ \\ \hline
        ~ & FedSA-GCL & $\mathbf{82.1}_{\pm 0.4}$ & $70.2_{\pm 0.4}$ & $\mathbf{83.9}_{\pm 0.9}$ & $88.8_{\pm 0.4}$ & $\mathbf{84.9}_{\pm 0.7}$ & $\mathbf{82.6}_{\pm 2.0}$ & $\mathbf{94.2}_{\pm 0.1}$ & $\mathbf{61.4}_{\pm 0.5}$ & $\mathbf{81.0}_{\pm 2.4}$ \\ \hline
    \multicolumn{11}{l}{\footnotesize \textit{Note:} The random label sparsity rate is set to 10\%.} \\
    \end{tabular}}
\end{table*}

\begin{table*}
    \centering
    \caption{Performance under Random Topology Sparsity}
    \label{table:topology sparsity}
    \resizebox{\textwidth}{!}{
    \begin{tabular}{c|c|ccccccccc}
    \hline
        \textbf{type} & \textbf{Method} & \textbf{Cora} & \textbf{CiteSeer} & \textbf{PubMed} & \textbf{CS} & \textbf{Photo} & \textbf{Computers} & \textbf{Physics} & \textbf{ogbn-arxiv} & \textbf{Overall} \\ \hline
        ~ & FedAvg & $65.0_{\pm 4.1}$ & $71.3_{\pm 0.3}$ & $82.2_{\pm 0.6}$ & $72.3_{\pm 5.9}$ & $75.4_{\pm 4.4}$ & $59.8_{\pm 5.5}$ & $88.0_{\pm 0.5}$ & $53.0_{\pm 1.0}$ & $70.9_{\pm 3.2}$ \\ 
        FL & FedProx & $60.0_{\pm 5.6}$ & $71.3_{\pm 0.3}$ & $82.2_{\pm 0.5}$ & $73.0_{\pm 3.3}$ & $74.0_{\pm 1.9}$ & $55.3_{\pm 3.0}$ & $87.9_{\pm 0.5}$ & $54.0_{\pm 1.1}$ & $69.7_{\pm 3.8}$ \\ 
        ~ & MOON & $62.0_{\pm 1.4}$ & $71.6_{\pm 0.7}$ & $71.0_{\pm 1.7}$ & $67.3_{\pm 0.8}$ & $74.6_{\pm 7.3}$ & $58.9_{\pm 2.8}$ & $87.3_{\pm 0.3}$ & $50.2_{\pm 1.6}$ & $67.9_{\pm 3.1}$ \\ \hline
        ~ & FedSage+ & $74.8_{\pm 1.7}$ & $69.9_{\pm 0.6}$ & $81.8_{\pm 1.7}$ & $79.0_{\pm 0.7}$ & $79.1_{\pm 2.0}$ & $69.8_{\pm 2.5}$ & $88.3_{\pm 0.3}$ & $55.8_{\pm 1.3}$ & $74.8_{\pm 2.2}$ \\ 
        FGL & FedGTA & $73.5_{\pm 2.5}$ & $71.4_{\pm 0.3}$ & $80.9_{\pm 1.2}$ & $\mathbf{89.5}_{\pm 0.2}$ & $83.6_{\pm 1.6}$ & $72.8_{\pm 1.2}$ & $\mathbf{94.1}_{\pm 0.1}$ & $59.1_{\pm 0.4}$ & $78.1_{\pm 2.9}$ \\ 
        ~ & FedTAD & $52.0_{\pm 8.6}$ & $\mathbf{71.7}_{\pm 0.5}$ & $71.6_{\pm 7.0}$ & $68.4_{\pm 2.5}$ & $73.4_{\pm 8.4}$ & $61.4_{\pm 4.3}$ & $88.4_{\pm 3.1}$ & $49.2_{\pm 2.3}$ & $67.0_{\pm 3.9}$ \\ \hline
        ~ & FedAsyn & $78.2_{\pm 0.5}$ & $71.0_{\pm 0.3}$ & $76.5_{\pm 0.8}$ & $78.4_{\pm 0.6}$ & $81.8_{\pm 1.2}$ & $74.4_{\pm 0.7}$ & $88.4_{\pm 0.3}$ & $46.8_{\pm 0.7}$ & $74.4_{\pm 1.8}$ \\ 
        AFL & TWAFL & $64.5_{\pm 1.5}$ & $47.1_{\pm 1.8}$ & $71.5_{\pm 0.4}$ & $69.3_{\pm 0.7}$ & $78.7_{\pm 1.7}$ & $66.3_{\pm 2.6}$ & $83.3_{\pm 0.8}$ & $38.6_{\pm 1.2}$ & $64.9_{\pm 3.8}$ \\ 
        ~ & FedBuff & $75.5_{\pm 0.5}$ & $70.6_{\pm 0.3}$ & $\mathbf{84.0}_{\pm 0.3}$ & $84.8_{\pm 2.0}$ & $77.4_{\pm 3.4}$ & $67.3_{\pm 1.8}$ & $88.2_{\pm 0.5}$ & $53.9_{\pm 0.5}$ & $75.2_{\pm 2.5}$ \\
        ~ & SWESALT & $73.0_{\pm 0.9}$ & $69.7_{\pm 0.5}$ & $81.0_{\pm 0.7}$ & $76.8_{\pm 1.0}$ & $79.7_{\pm 0.3}$ & $71.8_{\pm 0.5}$ & $87.8_{\pm 0.2}$ & $47.2_{\pm 0.7}$ & $73.4_{\pm 0.1}$ \\ \hline
        ~ & FedSA-GCL & $\mathbf{82.2}_{\pm 0.4}$ & $71.0_{\pm 0.4}$ & $83.9_{\pm 0.6}$ & $89.0_{\pm 0.2}$ & $\mathbf{85.2}_{\pm 1.0}$ & $\mathbf{82.0}_{\pm 3.3}$ & $94.0_{\pm 0.2}$ & $\mathbf{59.8}_{\pm 0.7}$ & $\mathbf{80.9}_{\pm 2.3}$ \\ \hline
    \multicolumn{11}{l}{\footnotesize \textit{Note:} The random topology sparsity rate is set to 10\%.} \\
    \end{tabular}}
\end{table*}

\begin{figure}
    \centering
    \includegraphics[width=1\linewidth]{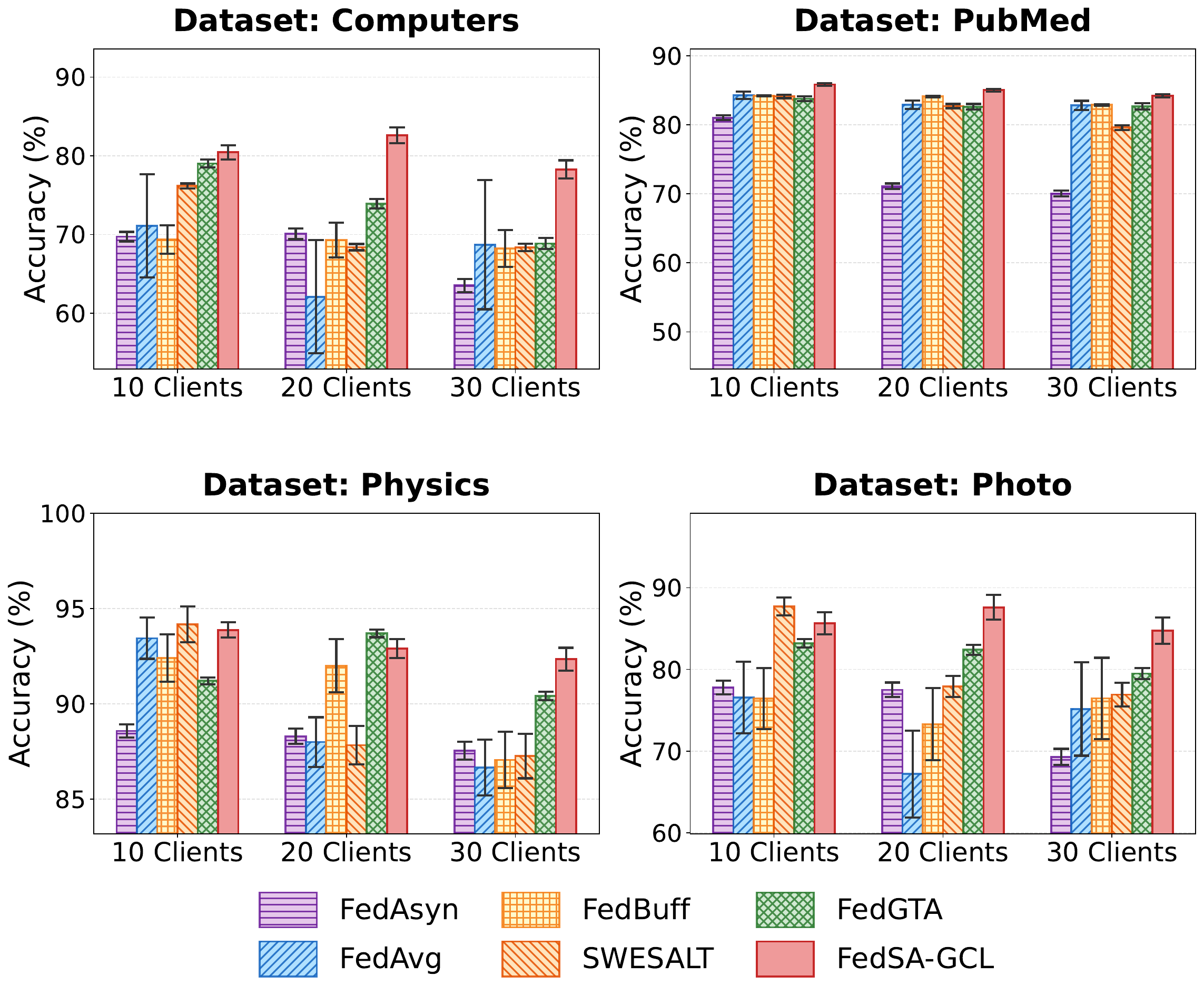}
    \caption{Performance comparison under varying numbers of participating clients.}
    \label{fig:client_num}
\end{figure}

To evaluate robustness under varying network scales, we partition the Computers, PubMed, Physics, and Photo datasets into 10, 20, and 30 clients using the Louvain algorithm. We benchmark FedSA-GCL against FedAsyn, FedAvg, FedBuff, FedGTA, and SWESALT, reporting the mean performance and 95\% confidence intervals across five independent runs. As Fig.~\ref{fig:client_num} illustrates, FedSA-GCL effectively mitigates performance degradation as the network expands, demonstrating consistent stability and adaptability across diverse client populations.

\subsection{Efficiency Analysis}
\label{sec:efficiency_analysis}

To comprehensively answer \textbf{Q5}, we evaluate the efficiency of the proposed FedSA-GCL framework across three distinct dimensions: empirical convergence speed (measured by client trips), absolute communication bandwidth overhead, and theoretical computational complexity.

\begin{figure}[htbp]
    \centering
    \subfigure[3D Visualization of Client Trips across on Louvain\label{fig:3D Louvain}]{
        \includegraphics[width=0.8\columnwidth]{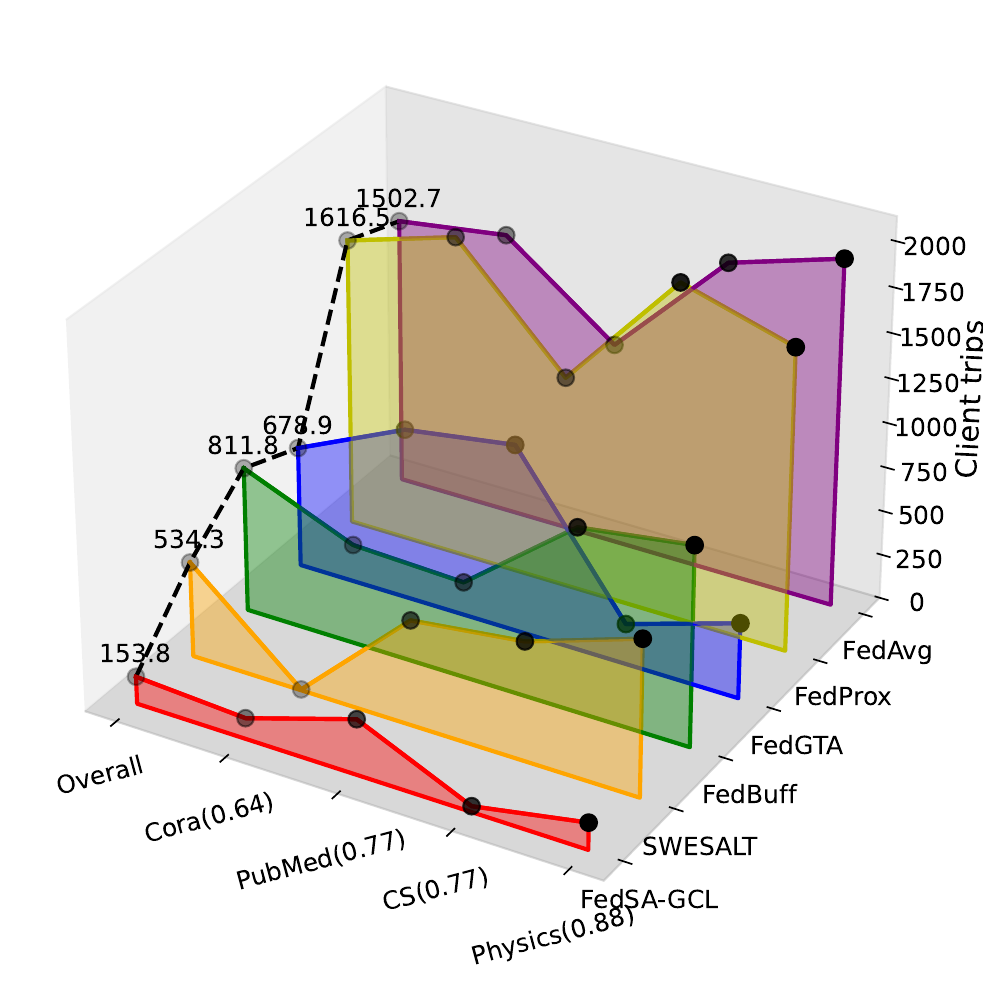}
    }
    \subfigure[3D Visualization of Client Trips across on Metis\label{fig:3D Metis}]{
        \includegraphics[width=0.8\columnwidth]{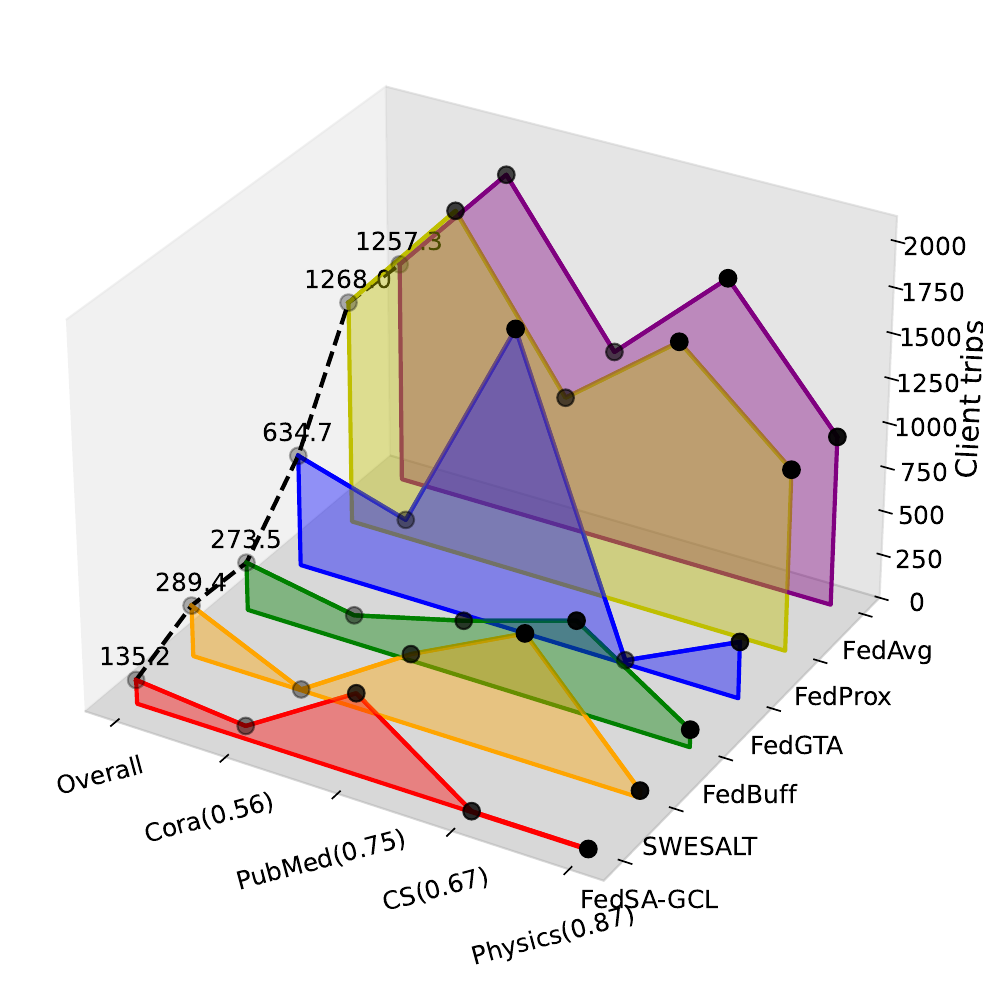}
    }
    \caption{3D Visualization of Client Trips across Methods.}
    \label{fig:3D}
\end{figure}

First, we follow the evaluation approach established in the semi-asynchronous algorithm FedBuff~\cite{nguyen2022federated} and adopt client trips as the evaluation metric for convergence speed. This metric represents the number of client-to-server communications required to reach a target validation accuracy. As it dynamically reflects both the communication and local computation phases, it serves as an effective proxy for overall training time. The 3D visualizations in Fig.~\ref{fig:3D} illustrate the client trips required by different baseline methods under Louvain and Metis partitioning strategies, respectively.

\begin{figure*}
    \centering
    \includegraphics[width=1\textwidth]{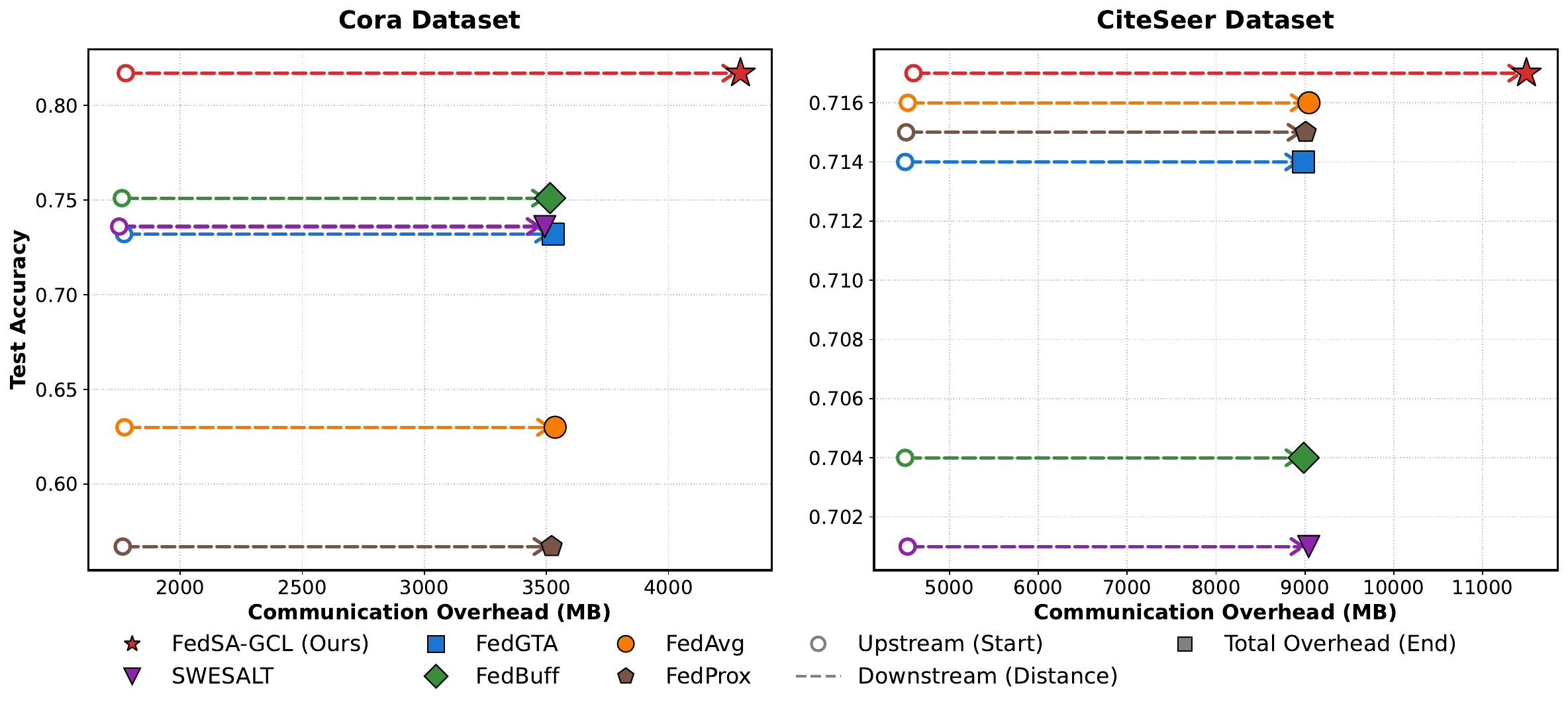}
    \caption{Total communication overhead breakdown against test accuracy.}
    \label{fig:comm_scatter}
\end{figure*}

To thoroughly analyze the absolute communication bandwidth required to complete the training process, we evaluate the cumulative upstream and downstream byte costs on the Cora and CiteSeer datasets (partitioned via Metis). 
Model weights constitute the primary transmission payload, whereas the auxiliary clustering metadata required by our method (i.e., LSC and SFM) occupy a negligible fraction of the bandwidth, amounting to only a few kilobytes. Because our evaluation standardizes execution based on client trips, the total number of client-to-server uploads remains mathematically fixed across all algorithms for a given number of evaluation steps. Consequently, as indicated by the starting markers in Fig.~\ref{fig:comm_scatter}, which explicitly denote the upstream overhead, the cumulative upstream overheads of all methods are strictly aligned.

The divergence in the total training communication overhead stems entirely from downstream transmissions, represented by the dashed lines in Fig.~\ref{fig:comm_scatter}. FedSA-GCL incurs a larger downstream overhead compared to the baselines. This is a direct consequence of the ClusterCast mechanism, which proactively broadcasts global updates to inactive clients within the same semantic cluster. Therefore, rather than claiming an absolute reduction in raw communication volume, FedSA-GCL demonstrates a highly cost-effective trade-off: by deliberately investing higher downstream bandwidth to proactively correct local semantic drift, it efficiently accelerates global convergence and achieves a significantly higher test accuracy (illustrated on the y-axis).

While the empirical results confirm that FedSA-GCL leverages a strategic communication trade-off to achieve superior accuracy, it is equally critical to ensure that these sophisticated mechanisms do not introduce prohibitive computational bottlenecks on local edge devices. Therefore, beyond communication efficiency, an analysis of the theoretical computational complexity is essential for evaluating the ultimate scalability of FGL methods.

Table~\ref{table:complexity} summarizes the memory and time complexities on both the client and server sides. Assuming a standard $L$-layer Graph Convolutional Network (GCN) backbone, we denote graph properties by nodes $n = |\mathcal{V}|$, edges $m = |\mathcal{E}|$, classes $c = |\mathcal{Y}|$, and feature dimensions $f$. Additional algorithmic parameters include feature propagation steps $k$, total participating clients $N$, asynchronous buffer size $K$, dynamic local epochs $E$ (for SWESALT), and augmented nodes $s$ with generated neighbors $g$ (for FedSage+).

For standard methods (e.g., FedAvg, FedAsyn), client-side time complexity is strictly bounded by GNN propagation at $\mathcal{O}(Lmf + Lnf^2)$, while regularization-based approaches (FedProx, MOON) merely introduce marginal additive terms for parameter penalties or contrastive losses. On the server side, asynchronous methods (FedAsyn, SWESALT) bypass the need to simultaneously cache all $N$ client models, effectively reducing the server memory requirement to $\mathcal{O}(Lf^2)$ or $\mathcal{O}(KLf^2)$, respectively.

\begin{table*}[htbp]
    
    \centering
    \caption{Algorithm complexity analysis for existing FGL optimization strategies.}
    \label{table:complexity}
    \resizebox{\textwidth}{!}{
    \begin{tabular}{lcccc}
    \toprule
    \textbf{Method} & \textbf{Client Mem.} & \textbf{Server Mem.} & \textbf{Client Time} & \textbf{Server Time} \\
    \midrule
    FedAvg & $\mathcal{O}(Lnf + Lf^2)$ & $\mathcal{O}(NLf^2)$ & $\mathcal{O}(Lmf + Lnf^2)$ & $\mathcal{O}(NLf^2)$ \\
    FedAsyn & $\mathcal{O}(Lnf + Lf^2)$ & $\mathcal{O}(Lf^2)$ & $\mathcal{O}(Lmf + Lnf^2)$ & $\mathcal{O}(Lf^2)$ \\
    FedProx & $\mathcal{O}(Lnf + 2Lf^2)$ & $\mathcal{O}(NLf^2)$ & $\mathcal{O}(Lmf + Lnf^2 + Lf^2)$ & $\mathcal{O}(NLf^2)$ \\
    MOON & $\mathcal{O}(Lnf + 3Lf^2)$ & $\mathcal{O}(NLf^2)$ & $\mathcal{O}(Lmf + Lnf^2 + nf)$ & $\mathcal{O}(NLf^2)$ \\
    FedProto & $\mathcal{O}(Lnf + Lf^2 + c)$ & $\mathcal{O}(Nc)$ & $\mathcal{O}(Lmf + Lnf^2 + nc)$ & $\mathcal{O}(Nc)$ \\
    FedSage+ & $\mathcal{O}(L(n+sg)f + 3Lf^2)$ & $\mathcal{O}(3NLf^2)$ & $\mathcal{O}(L(m+sg)f + L(n+sg)f^2)$ & $\mathcal{O}(NLf^2)$ \\
    FedGTA & $\mathcal{O}(Lnf + Lf^2 + c)$ & $\mathcal{O}(NLf^2 + Nkc)$ & $\mathcal{O}(Lmf + Lnf^2 + k^2nc)$ & $\mathcal{O}(NLf^2 + Nkc)$ \\
    SWESALT & $\mathcal{O}(Lnf + Lf^2)$ & $\mathcal{O}(KLf^2)$ & $\mathcal{O}(E(Lmf + Lnf^2) + Lf^2)$ & $\mathcal{O}(KLf^2)$ \\
    \midrule
    \textbf{FedSA-GCL} & $\mathcal{O}(Lnf + Lf^2 + c^2)$ & $\mathcal{O}(NLf^2 + Nc^2)$ & $\mathcal{O}(Lmf + Lnf^2 + kmc + nc^2)$ & $\mathcal{O}(NLf^2 + N^2c^2)$ \\
    \bottomrule
    \end{tabular}}

\end{table*}

In FedSA-GCL, the additional client-side computation stems from generating the SFM and calculating the LSC. By leveraging the matrix formulation, SFM is computed via sparse-dense matrix multiplications as $\hat{\mathbf{Y}}_i^\top \tilde{\mathbf{A}}_i \hat{\mathbf{Y}}_i$, which takes $\mathcal{O}(mc + nc^2)$ time. Simultaneously, the LSC calculation involves a fixed $k$-step Non-param LP, adding $\mathcal{O}(kmc)$ to the time complexity. Since the number of classes $c$ and propagation steps $k$ are small constants, the total additional overhead $\mathcal{O}(kmc + nc^2)$ remains strictly bounded by the GNN backbone's complexity. Meanwhile, the server-side similarity calculation takes $\mathcal{O}(N^2c^2)$ time, which remains highly manageable due to the extremely small dimension of $c$.

A key advantage of this computational design becomes apparent when scaling to dense graph topologies, where the number of edges $m$ can approach $\mathcal{O}(n^2)$. As shown in Table~\ref{table:complexity}, the additional client time complexity introduced by FedSA-GCL scales strictly linearly with $m$ ($\mathcal{O}(kmc + nc^2)$). In practice, the number of classes $c$ and propagation steps $k$ are small constants compared to the feature dimension $f$ ($c, k \ll f$). Consequently, thanks to the sparse matrix optimizations, the added overhead remains strictly bounded by the baseline message-passing complexity, $\mathcal{O}(Lmf)$. By carefully avoiding any $\mathcal{O}(m^2)$ operations, FedSA-GCL ensures that its computational and memory requirements remain highly efficient, even as graph density increases.

\section{Conclusion}
\label{Conclusion}

In this paper, we proposed FedSA-GCL, a novel semi-asynchronous FGL framework designed to address the unique challenges posed by non-IID graph-structured data under asynchronous settings. Our method incorporates a staleness-aware LSC aggregation strategy, which effectively mitigates the adverse impact of stale models and accelerates convergence. Additionally, we introduced ClusterCast, a cluster-level broadcasting mechanism that enables inactive but semantically similar clients to benefit from timely model updates, thereby enhancing cross-client consistency. Furthermore, we designed a clustering algorithm based on the SFM to capture semantic similarities and demonstrate superior adaptability across training rounds. 
Crucially, we provide a formal convergence analysis to theoretically guarantee the optimization stability of the proposed framework.

Extensive experiments on eight real-world graph datasets, scaling up to the large \texttt{ogbn-arxiv} network, including ablation studies, robustness tests, hyperparameter sensitivity, and training efficiency evaluations, confirm that FedSA-GCL consistently outperforms state-of-the-art baselines in terms of accuracy, robustness, and convergence speed. Notably, the algorithm circumvents quadratic edge operations, guaranteeing exceptional computational scalability even on highly dense graph topologies. Particularly, our method demonstrates strong resilience to label and topology sparsity, and significantly accelerates the global optimization process, drastically reducing the total computational delays and the required number of client trips. This high convergence efficiency stems from a strategic system trade-off: by intentionally investing slightly more downstream bandwidth to proactively correct local semantic drift, FedSA-GCL successfully circumvents the severe performance degradation typical of asynchronous non-IID environments. These results collectively validate the effectiveness and practicality of FedSA-GCL in real-world federated graph learning scenarios. Future work will explore extending this framework to dynamic graphs where node features and topologies evolve over time, further enhancing its applicability in non-stationary environments.

\section*{Declaration of Competing Interest}
The authors declare that they have no known competing financial interests or personal relationships that could have appeared to influence the work reported in this paper.

\section*{Acknowledgements}
This research was supported by Shenzhen Fundamental Research Program (JCYJ20230807094104009).

\section*{Data availability}
Data will be made available on request.

\bibliographystyle{elsarticle-num}

\bibliography{references}



\end{document}